\documentclass[11pt,a4paper]{article}
\usepackage[margin=22mm]{geometry}
\usepackage[T1]{fontenc}
\usepackage[utf8]{inputenc}
\usepackage{lmodern}
\usepackage{amsmath,amssymb,textcomp}
\usepackage{graphicx}
\usepackage{booktabs,array}
\usepackage[font=small,labelfont=bf,labelsep=period]{caption}
\usepackage[super,sort&compress,comma]{natbib}
\usepackage{xurl}
\usepackage[hidelinks,unicode]{hyperref}
\hypersetup{pdftitle={ReTRACE: Feedback-Guided LLM Agent in a Chemistry Harness for Retrosynthesis Planning},pdfauthor={Jinxiang Xie, Yijian Zhang, Qingyuan Yi, Jianyi Wang, Hao Dong}}
\title{\LARGE\bfseries ReTRACE: Feedback-Guided LLM Agent in a Chemistry Harness for Retrosynthesis Planning}
\author{\parbox{\dimexpr\textwidth-2\tabcolsep\relax}{\centering\normalsize
Jinxiang Xie\textsuperscript{1,\textdagger}, Yijian Zhang\textsuperscript{1,\textdagger}, Qingyuan Yi\textsuperscript{2}, Jianyi Wang\textsuperscript{2,*}, Hao Dong\textsuperscript{1,*}\\[6pt]
{\small \textsuperscript{1}State Key Laboratory of Analytical Chemistry for Life Science, Kuang Yaming Honors School, Chemistry and Biomedicine Innovation Center (ChemBIC), ChemBioMed Interdisciplinary Research Center at Nanjing University, Nanjing University, Nanjing 210023, China;\\[4pt]
\textsuperscript{2}School of Medicine, Guangxi University, Nanning, 530004, China}
}}
\date{}
\begin{document}
\maketitle
\begingroup
\renewcommand{\thefootnote}{}
\footnotetext{\textsuperscript{\textdagger}These authors contributed equally to this work. \textsuperscript{*}Corresponding authors: Hao Dong and Jianyi Wang. E-mail: \href{mailto:donghao@nju.edu.cn}{donghao@nju.edu.cn} (Hao Dong); \href{mailto:jianyiwang@gxu.edu.cn}{jianyiwang@gxu.edu.cn} (Jianyi Wang).}
\endgroup
\begin{abstract}

Computer-aided retrosynthesis has advanced through single-step prediction, tree search and route scoring, but complex targets still require judgement over the whole route, where stereochemistry, functional-group compatibility, cost and greenness interact. Here we present \textbf{\mbox{ReTRACE}} (\textbf{RE}trosynthesis with \textbf{T}ool-orchestrated \textbf{R}easoning \textbf{A}nd \textbf{C}ontext-aware \textbf{E}vidence), an agent that places a large language model (LLM) in a chemistry harness that integrates molecular tools, complementary single-step prediction engines, a purchasable starting-material library, a route memory and a multi-criterion route evaluator. Routes are judged by a new metric that combines programmatic checks with adversarial LLM review, and its verdicts are compared with a synthetic chemist's. The agent learns continually from evaluator feedback during search, diagnosing weak routes and backtracking to revise them. With Claude Opus 4.8, at least one of the top 10 routes proposed by ReTRACE passed the audit for 87.4\%, 95.0\% and 86.7\% of targets on the three benchmarks. Two prospective wet-lab syntheses show that ReTRACE route proposals can be converted into executable procedures after expert revision.

\end{abstract}


\section{Introduction}\label{introduction}

Retrosynthetic analysis is one of the central tasks of organic synthesis. Since Corey formalized the logic of working backwards from a target molecule to simpler precursors,\cite{ref1,ref2,ref3} route design has been treated as a strategic activity that depends on chemical judgement about which bonds to break, in what order and under what conditions. This judgement matters increasingly in drug discovery, where complex candidates often contain several heterocycles, stereocentres and sensitive functional groups. For such targets, an early retrosynthetic choice propagates through the whole route and affects selectivity, protecting-group burden, cost, safety and sustainability.

Computer-aided synthesis planning (CASP) has progressed substantially since early expert systems encoded reaction rules and retrosynthetic templates as searchable chemical knowledge.\cite{ref1,ref2,ref4,ref5} Current data-driven systems couple single-step retrosynthesis models with purchasable-compound libraries and tree search.

Single-step models fall into two broad families. Template-based models apply reaction rules extracted from known reactions and so remain chemically traceable,\cite{ref6,ref7} whereas template-free models learn the product-to-precursor mapping directly from reaction data and can propose transformations outside any template library.\cite{ref8,ref9,ref10,ref11,ref12,ref13} Combining models with different inductive biases improves single-step prediction further.\cite{ref14} These advances are essential, but a single-step model judges one disconnection at a time and cannot by itself tell whether that disconnection serves the route as a whole.

Search algorithms then assemble these single-step predictions into multistep routes. Examples include the modular ASKCOS platform,\cite{ref15,ref16} Monte Carlo tree search guided by a neural template policy\cite{ref6,ref7} and AND--OR graph search with a learned value function in Retro*.\cite{ref17} Hybrid platforms extend the same framework to enzymatic steps.\cite{ref18}

Plausible single-step disconnections are often plentiful, and the harder decision is which precursor to expand for a given target. Chemists make this decision by weighing mechanism, functional-group compatibility, precedent, availability and cost together, and these signals rarely reduce to a single number. A highly ranked Suzuki--Miyaura disconnection, for example, may be unattractive if the required organoboron partner is unstable or expensive. Existing search controllers can explore very large trees, but they usually compress this judgement into model probabilities, value estimates or a scalar route score.\cite{ref6,ref7,ref17,ref19,ref20} This makes search fast, but it is hard to apply route-wide constraints or to see why a branch was chosen.

Large language models (LLMs) offer another way to make these decisions. Recent systems use them to guide route search,\cite{ref21} to coordinate single-step tools through reinforcement learning,\cite{ref22} to check route constraints during search\cite{ref23} and to evaluate synthetic strategy in natural language.\cite{ref24} These systems build on chemistry agents in which the LLM selects and interprets expert tools while deterministic software supplies verifiable observations.\cite{ref25,ref26} Their results suggest that LLMs are most useful for strategy assessment, constraint reasoning and tool coordination, and that specialized single-step models remain necessary for proposing precursors.

Recent gains in LLM capability create new opportunities for coordinating heterogeneous chemistry tools over long-horizon route planning. In software engineering, LLM agents now call tools over many rounds and revise their actions from execution feedback,\cite{ref27} and the length of tasks they can complete has grown rapidly.\cite{ref28} Chemistry poses an additional problem. A general-purpose LLM can discuss functional groups and synthetic strategy, but on its own it is unreliable at precise molecular operations such as writing valid SMILES, conserving atoms or computing quantitative scores. For long-horizon route planning, the environment built around the model, which we call its harness, therefore matters alongside the model itself. The harness determines which tools the model can call, what it remembers and what feedback it receives.

\begin{figure}[!t]
\centering
\includegraphics[width=\textwidth]{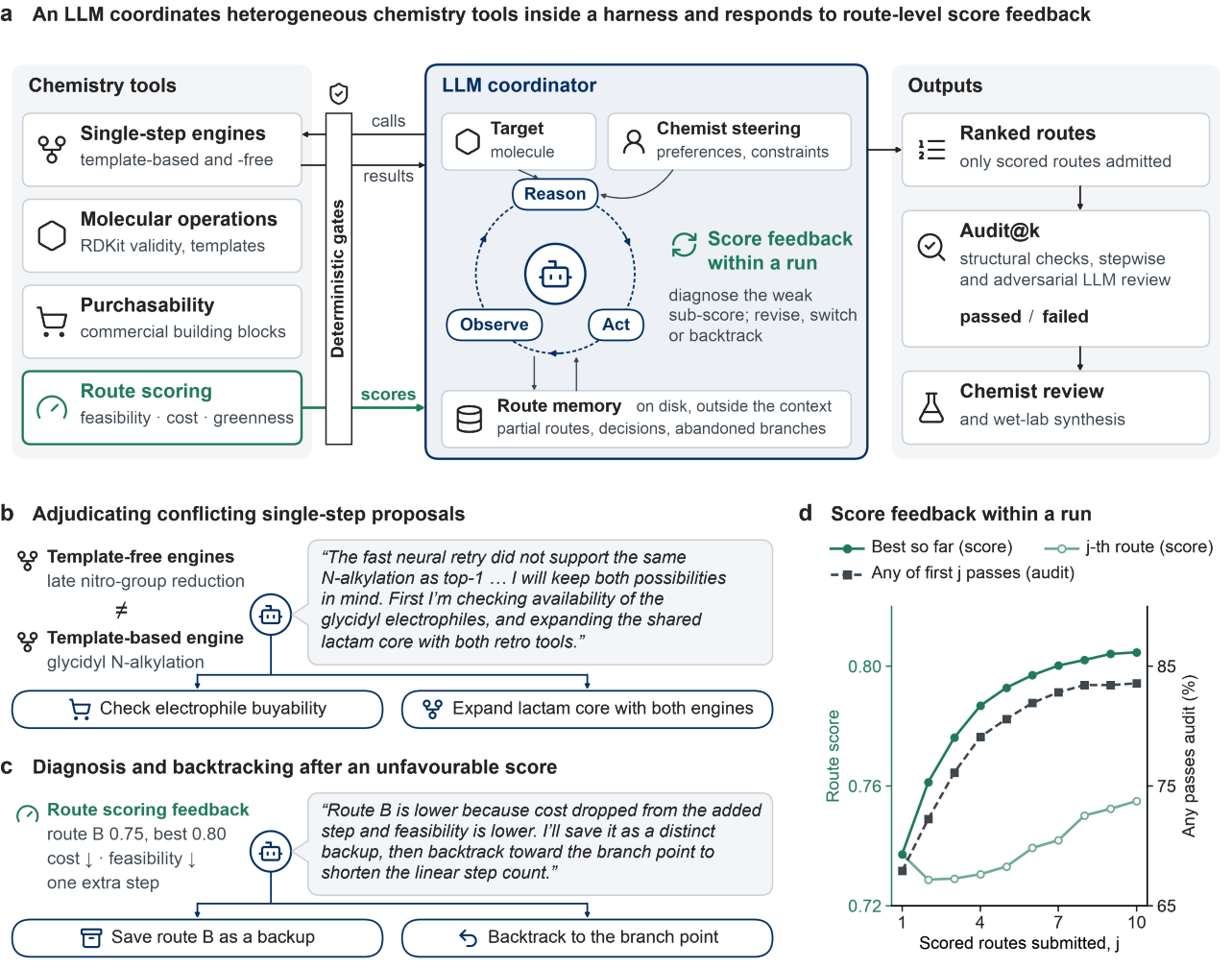}
\caption{\textbf{ReTRACE harness, two decision instances from real search trajectories and the response to score feedback within a run.} \textbf{a}, An LLM coordinator works inside a harness of heterogeneous chemistry tools, deterministic gates that check every tool call, tool result and saved route, and a route memory that carries partial routes, past decisions and abandoned branches into the next round. Route scores return to the coordinator as feedback within a run. Ranked routes are judged by Audit@k as passed or failed, and selected routes go to chemist review and wet-lab synthesis. \textbf{b}, Adjudicating conflicting single-step suggestions. The coordinator keeps both a late-stage nitro reduction and a glycidyl N-alkylation open, checks whether the electrophile can be purchased and expands the shared lactam core with both engine classes before choosing. \textbf{c}, Diagnosis and backtracking. Route B scores 0.75 against 0.80 for the best route. The coordinator attributes the gap to lower cost and feasibility from the added steps, saves route B as an alternative and backtracks to shorten the linear sequence. Monologues in b and c are verbatim excerpts from run trajectories. \textbf{d}, Score feedback within a run in 670 completed Retro*-190 runs. Mean score of the best route so far (dark green) and of the j-th scored route (light green; left axis), and percentage of runs in which at least one of the first j scored routes passed the audit (grey, dashed; right axis).}
\label{fig:1}
\end{figure}

We developed \textbf{\mbox{ReTRACE}} (\textbf{RE}trosynthesis with \textbf{T}ool-orchestrated \textbf{R}easoning \textbf{A}nd \textbf{C}ontext-aware \textbf{E}vidence) by building such a harness around an LLM for multistep retrosynthesis (Figure~\ref{fig:1}a). The harness has three parts. The tools supply what the model cannot do reliably itself: RDKit molecular operations, template-based and template-free single-step engines, purchasability lookup and route scoring, all called through deterministic gates that reject invalid structures in code. The route memory sits outside the context window and keeps partial routes, past decisions and abandoned branches across hours of search. The feedback comes from a route evaluator that scores feasibility, cost and greenness. Final routes are judged separately by a new metric that pairs programmatic structural checks with step-by-step and adversarial LLM review, and we compare this audit with the judgements of a synthetic chemist.

Within this harness the agent learns from feedback on two time scales. During a run, the evaluator's sub-scores show where a route is weak, and the agent uses them to backtrack to a specific step or to change its disconnection strategy (Figure~\ref{fig:1}c). As a run submits more scored routes, the best score so far and the chance that at least one route passes the audit both rise and level off after about eight routes (Figure~\ref{fig:1}d). Across runs, the harness records states, actions, tool returns and outcomes as trajectories that can be used to update the model itself.

ReTRACE is designed as an assistant that leaves experimental judgement to chemists. Users can steer it in natural language towards cheaper, greener or better-precedented routes. Because the route memory persists on disk, the sessions can be resumed, chemists can supply feedback or constraints, and the agent can continue planning with a human in the loop. Because the harness is modular, new expert tools, such as enzymatic single-step predictors, can be added without changing the coordinator.

This work makes four contributions. First, a large-scale audit of single-step engines shows that they contribute complementary candidates, in line with recent evidence for model ensembling.\cite{ref14} This finding motivates a harness that keeps heterogeneous evidence and lets the LLM weigh it at route level. Second, we build that harness and compare ReTRACE with classical and LLM controllers on three route-level benchmarks under a new metric whose verdicts we check against a synthetic chemist. Third, we examine how the agent learns from evaluator feedback, within a run through decision trajectories and a tool ablation, and across runs through a first attempt to train the model on harness trajectories. Fourth, two prospective wet-lab syntheses test whether ReTRACE routes can be turned into executable procedures after expert revision.

\section{Results}\label{results}

\subsection{Heterogeneous single-step tools provide complementary but imperfect candidate evidence}\label{heterogeneous-single-step-tools-provide-complementary-but-imperfect-candidate-evidence}

\begin{figure}[!t]
\centering
\includegraphics[width=\textwidth]{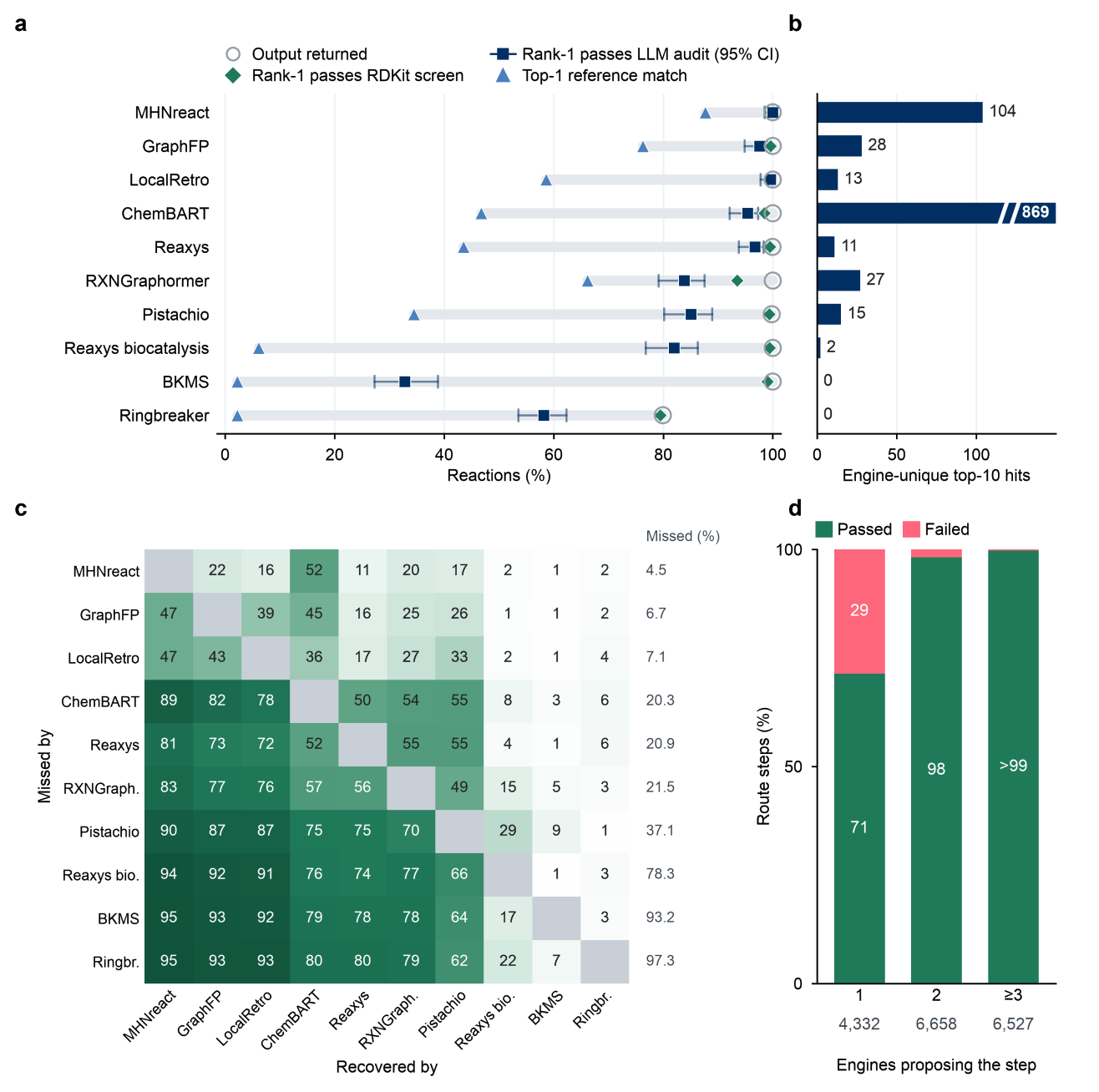}
\caption{\textbf{Validity, reference recovery and complementarity of ten single-step engines, and engine agreement in planned routes.} a--c, 50,000 USPTO-50K reactions, with engines ordered by top-10 recovery and reference matching. \textbf{a}, Fraction of reactions with output, with a rank-1 prediction that passes RDKit screening, with a rank-1 prediction that passes the LLM audit (95\% Wilson interval), and with an exact top-1 reference match. \textbf{b}, Reactions recovered within the top 10 by that engine alone. \textbf{c}, Percentage of the top-10 misses of the row engine that the column engine recovers within its top 10, and the right column gives the top-10 miss rate of each engine. \textbf{d}, Route steps in the Retro*-190 runs that passed or failed the audit, by the number of runtime engines that proposed the step.}
\label{fig:2}
\end{figure}

Figure~\ref{fig:2} compares ten single-step retrosynthesis engines on the 50,000 USPTO-50K reactions we recorded: GraphFP, RXNGraphormer, ChemBART, Reaxys template relevance, Ringbreaker, Pistachio, Reaxys biocatalysis, BKMS, MHNreact\cite{ref29} and LocalRetro\cite{ref30}. For each engine we asked four questions: whether it returned output, whether its rank-1 prediction passed RDKit structural screening, whether that prediction passed an adversarial LLM audit, and whether its predictions reproduced the reference precursors recorded in the dataset, ignoring stereochemistry. Most targets admit several valid disconnections, so exact match is a conservative measure of coverage and says little about chemical feasibility. The measures diverge for every engine (Figure~\ref{fig:2}a): rank-1 predictions pass the audit far more often than they reproduce the reference, and a single top-k accuracy is therefore a poor summary of an engine's value for route planning.

The engines are strongly complementary. The top-10 union of all ten engines recovers the reference precursors for 98.6\% of reactions. Even the misses of the strongest engines are mostly recovered elsewhere: 69.8\% of the top-10 misses of MHNreact, 79.6\% of those of GraphFP and 80.9\% of those of LocalRetro are recovered by at least one other engine. An engine's contribution to the pool does not follow its own ranking. ChemBART is fourth on top-10 recovery, yet it accounts for most of the reactions that no other engine hits (Figure~\ref{fig:2}b) and recovers the largest share of the misses of MHNreact (Figure~\ref{fig:2}c). These observations agree with the ensembling results of RetroChimera at the single-step level\cite{ref14} and motivate our design choice for the agent system, which retains heterogeneous candidates and lets the coordinator select and backtrack in the context of the full route, rather than fixing one single-step predictor before search begins.

Agreement between engines also carries information during multistep planning. We traced each step of the audited top-10 routes in the four Retro*-190 runs (Table~\ref{tab:performance}) back to the candidates that the four runtime engines (GraphFP, RXNGraphormer, ChemBART and Ringbreaker) had returned in the same run, matching molecules without stereochemistry and excluding steps that no engine had proposed. Steps proposed by two or more engines almost always passed the audit (98.2\% with two engines and 99.8\% with three or more), whereas 28.6\% of steps proposed by a single engine failed (Figure~\ref{fig:2}d). Agreement between engines is therefore a useful signal of step reliability when the coordinator weighs candidates.

The error modes of the engines also differ. Among returned rank-1 predictions, the two generative engines, ChemBART and RXNGraphormer, had the lowest structural pass rates, and their failures were more often unparsable structures, atom or valence anomalies and mechanistically untenable precursor combinations. All engines were run with their released or historical checkpoints, and the recorded corpus overlaps their training data, so Figure~\ref{fig:2} describes retrospective performance and does not rank the engines on held-out data. ReTRACE therefore does not adopt the raw ranking of any engine. It canonicalizes, de-duplicates and screens candidates before exposing them to the LLM.

\subsection{Route-level performance and controller comparison}\label{route-level-performance-and-controller-comparison}

ReTRACE converts single-step candidates into complete multistep synthesis plans. With the LLM coordinators, run success, that is, the fraction of targets for which a run completes and returns at least one route, was 93.3--100\% across the three datasets (Table~\ref{tab:performance}). Run success, however, does not imply chemical validity: many of these routes contain steps that would not work in the laboratory, partly because single-step models can return false-positive precursors. Route-level metrics widely used in related work, such as solve rates based on reaching purchasable leaves, are therefore not sufficient to assess our system. We instead propose a new metric, Audit@k, which applies the same frozen two-stage audit to every method and records whether at least one of the top-k ranked routes passes the audit, with all targets in the dataset as the denominator. Failed runs and targets without output count as failures. On Retro*-190, Audit@1/3/5/10 were 70.0\%/78.4\%/83.7\%/87.4\% for Claude Opus 4.8 and 65.8\%/75.3\%/79.5\%/82.6\% for GPT-5.6 Sol standard. The same systems also performed well on JMC2025 and FDA30, and Table~\ref{tab:performance} gives all point estimates and run success.

The coordinating model and its search preset affect route-planning outcomes. ReTRACE provides several reasoning and search budgets to suit different users, but larger reasoning and search budgets did not yield monotonic gains. For GPT-5.6 Sol fast, standard and thorough, Audit@10 was 80.0\%, 82.6\% and 82.1\% on Retro*-190, 89.4\%, 92.2\% and 88.7\% on JMC2025, and 86.7\%, 90.0\% and 83.3\% on FDA30. The standard preset gave the highest Audit@1 and Audit@10 of the three presets on every dataset. The thorough preset, which has the largest reasoning effort, query budget and route capacity, never exceeded it and was at or below the fast preset in five of the six comparisons. More reasoning and search on a target therefore did not translate into proportionally better routes.

For comparison, the classical Retro* search baseline achieved Audit@1 of 73.7\%, 80.9\% and 53.3\% on the three datasets, and the general-purpose agent baseline Codex with GPT-5.6 Sol standard, which generates routes directly, achieved 32.1\%, 40.4\% and 30.0\%. These external baselines share the target sets, the stock-termination criterion and the frozen audit, but keep their native controllers and budgets, so the comparison is outcome-level rather than compute-matched. Retro* achieved the highest first-route coverage on Retro*-190 and was second only to Claude Opus 4.8 on JMC2025 (80.9\% versus 82.3\%), but it returns only one route per target. ReTRACE's multi-candidate sets increase coverage as k grows and reach higher Audit@10 on all three datasets.

To test the agreement of this audit endpoint with human chemical judgement, a researcher with a background in synthetic chemistry reviewed 59 candidate routes returned for the first six Retro*-190 targets, including all 24 rank-1 routes under four coordinator conditions and selected lower-ranked candidates, comprising 274 reaction steps. Under the same audit-pass/audit-fail criterion used for Audit@k, route-level agreement between the automatic audit and the chemist was 96.6\% (57/59; Cohen's \ensuremath{\kappa} = 0.93), rank-1 agreement was 95.8\% (23/24; \ensuremath{\kappa} = 0.91) and step-level agreement was 97.4\% (267/274; \ensuremath{\kappa} = 0.89). Only two route-level judgements disagreed, one in each direction.

\begin{table}[p]
\centering
\caption{Route-planning performance of coordinating models and unified audit baselines across three datasets, with Claude Opus 4.8 tool-group ablations on Retro*-190.}
\label{tab:performance}
\fontsize{8.5}{10.5}\selectfont
\setlength{\tabcolsep}{2.2pt}
\renewcommand{\arraystretch}{1.15}
\begin{tabular}{@{}>{\raggedright\arraybackslash}m{42mm}>{\centering\arraybackslash}m{14mm}*{8}{>{\centering\arraybackslash}m{11mm}}@{}}
\toprule[0.9pt]
\textbf{Controller / baseline} & \shortstack{\textbf{Run}\\\textbf{success}\\\textbf{(\%)}} &
\multicolumn{4}{c}{\begin{tabular}[c]{@{}*{4}{>{\centering\arraybackslash}m{11mm}}@{}}
\multicolumn{4}{c}{\parbox[c][10mm][c]{48mm}{\centering\textbf{Audit@$k$ (\%)}}} \\
\cmidrule(lr){1-4}
$k=1$ & $k=3$ & $k=5$ & $k=10$
\end{tabular}} &
\multicolumn{4}{c}{\begin{tabular}[c]{@{}*{4}{>{\centering\arraybackslash}m{11mm}}@{}}
\multicolumn{4}{c}{\parbox[c][10mm][c]{48mm}{\centering\textbf{Rank-1 route length}\\\textbf{vs reference (\%)}}} \\
\cmidrule(lr){1-4}
Shorter & Equal & Longer & None
\end{tabular}} \\
\specialrule{0.9pt}{4pt}{5pt}
\multicolumn{10}{@{}l}{\textbf{Retro*-190} (N = 190)} \\
\addlinespace[2pt]
Claude Opus 4.8 & 97.9 & 70.0 & \textbf{78.4} & \textbf{83.7} & \textbf{87.4} & 65.8 & 15.8 & 18.4 & 0.0 \\
GPT-5.6 Sol fast & 100.0 & 64.2 & 76.8 & 78.4 & 80.0 & 64.2 & 16.3 & 19.5 & 0.0 \\
GPT-5.6 Sol standard & 100.0 & 65.8 & 75.3 & 79.5 & 82.6 & 64.7 & 11.6 & 23.7 & 0.0 \\
GPT-5.6 Sol thorough & 100.0 & 63.7 & 72.6 & 79.5 & 82.1 & 60.0 & 15.8 & 24.2 & 0.0 \\
Qwen3.6-35B-A3B & 91.6 & 23.7 & 33.2 & 33.7 & 34.7 & 74.7 & 12.1 & 8.4 & 4.7 \\
\specialrule{0.25pt}{3pt}{3pt}
Retro* & 85.8 & \textbf{73.7} & 73.7 & 73.7 & 73.7 & 42.6 & 21.6 & 20.5 & 15.3 \\
Codex & 98.9 & 32.1 & 32.1 & 32.1 & 32.1 & 57.9 & 24.7 & 16.3 & 1.1 \\
\specialrule{0.25pt}{3pt}{3pt}
w/o single-step prediction & 98.4 & 42.3 & 57.7 & 68.8 & 77.8 & 67.7 & 12.2 & 20.1 & 0.0 \\
w/o route scoring & 99.5 & 52.1 & 76.3 & 81.6 & 86.8 & 65.3 & 14.7 & 20.0 & 0.0 \\
w/o molecular-perception tools & 100.0 & 60.5 & 73.2 & 76.8 & 84.2 & 68.4 & 15.3 & 16.3 & 0.0 \\
\specialrule{1pt}{7pt}{5pt}
\multicolumn{10}{@{}l}{\textbf{JMC2025} (N = 141)} \\
\addlinespace[2pt]
Claude Opus 4.8 & 99.3 & \textbf{82.3} & \textbf{87.9} & \textbf{90.1} & \textbf{95.0} & 63.8 & 14.9 & 21.3 & 0.0 \\
GPT-5.6 Sol fast & 100.0 & 75.2 & 85.8 & 88.7 & 89.4 & 60.3 & 12.8 & 27.0 & 0.0 \\
GPT-5.6 Sol standard & 100.0 & 75.9 & 85.8 & 88.7 & 92.2 & 66.7 & 13.5 & 19.9 & 0.0 \\
GPT-5.6 Sol thorough & 100.0 & 75.2 & 85.1 & 85.8 & 88.7 & 62.4 & 12.8 & 24.8 & 0.0 \\
Qwen3.6-35B-A3B & 92.2 & 24.8 & 39.7 & 42.6 & 43.3 & 68.1 & 7.8 & 19.1 & 5.0 \\
\specialrule{0.25pt}{3pt}{3pt}
Retro* & 95.0 & 80.9 & 80.9 & 80.9 & 80.9 & 44.7 & 15.6 & 34.8 & 5.0 \\
Codex & 100.0 & 40.4 & 40.4 & 40.4 & 40.4 & 58.2 & 12.8 & 27.7 & 1.4 \\
\specialrule{1pt}{7pt}{5pt}
\multicolumn{10}{@{}l}{\textbf{FDA30} (N = 30)} \\
\addlinespace[2pt]
Claude Opus 4.8 & 93.3 & 70.0 & 80.0 & 83.3 & 86.7 & 73.3 & 13.3 & 10.0 & 3.3 \\
GPT-5.6 Sol fast & 96.7 & 70.0 & 80.0 & 83.3 & 86.7 & 73.3 & 20.0 & 6.7 & 0.0 \\
GPT-5.6 Sol standard & 100.0 & \textbf{76.7} & \textbf{83.3} & \textbf{86.7} & \textbf{90.0} & 80.0 & 13.3 & 6.7 & 0.0 \\
GPT-5.6 Sol thorough & 100.0 & 66.7 & 76.7 & 76.7 & 83.3 & 76.7 & 10.0 & 13.3 & 0.0 \\
Qwen3.6-35B-A3B & 100.0 & 30.0 & 53.3 & 60.0 & 63.3 & 83.3 & 13.3 & 3.3 & 0.0 \\
\specialrule{0.25pt}{3pt}{3pt}
Retro* & 86.7 & 53.3 & 53.3 & 53.3 & 53.3 & 36.7 & 16.7 & 13.3 & 33.3 \\
Codex & 100.0 & 30.0 & 30.0 & 30.0 & 30.0 & 66.7 & 10.0 & 20.0 & 3.3 \\
\bottomrule[0.9pt]
\end{tabular}
\par\vspace{6pt}
\begin{minipage}{\textwidth}
\fontsize{9}{11}\selectfont
Within each dataset, rows above the first thin rule are ReTRACE with different coordinating models and presets, and rows below it are external baselines. The three rows below the second thin rule of the Retro*-190 block are tool-group ablations of Claude Opus 4.8 (w/o, without), each run once with one tool group removed and compared target by target with the full Claude Opus 4.8 row. Bold marks the highest Audit@k within each dataset. Run success is the percentage of targets whose run completed and returned at least one route. Audit@k is the percentage of all targets with at least one audit-passing route among the top k candidates. Rank-1 route length compares the number of reaction steps in the rank-1 route with the reference route, and the shorter, equal, longer and no-route shares sum to 100\%. No route means that no rank-1 route with at least one reaction step was saved for the target. This happens when a run ends without saving a complete route. Runs that saved routes before ending with an error count as failed runs for run success and Audit@k, but their rank-1 route is still classified by length, so the no-route share can be lower than the share of failed runs.
\end{minipage}
\end{table}

The run trajectories show how the coordinator uses this feedback within a run (Figure~\ref{fig:1}d). We analysed the 670 completed Retro*-190 runs (Claude Opus 4.8 and GPT-5.6 Sol fast, standard and thorough) in which every scored route was audited. As a run submits more scored routes, the best score so far and the percentage of runs with at least one passed route both rise and level off after about eight routes. The score of the j-th route is more informative. Runs submit 3--10 scored routes, so this curve averages fewer runs at larger j (670 runs for j \ensuremath{\leq} 3 and 282 for j = 10).

\subsection{Route-level chemical judgement in a representative decision trajectory}\label{route-level-chemical-judgement-in-a-representative-decision-trajectory}

Quantitative Audit@k only shows whether a candidate set contains an audit-passing route; it does not show how the coordinator handles conflicting single-step evidence. To make this decision process inspectable, we follow one real execution trajectory, presenting the target, tool returns, coordinator diagnosis, next tool call and final route, and compare rejected branches with the reference or literature route in the same chemical context.

\begin{figure}[p]
\centering
\includegraphics[width=\textwidth]{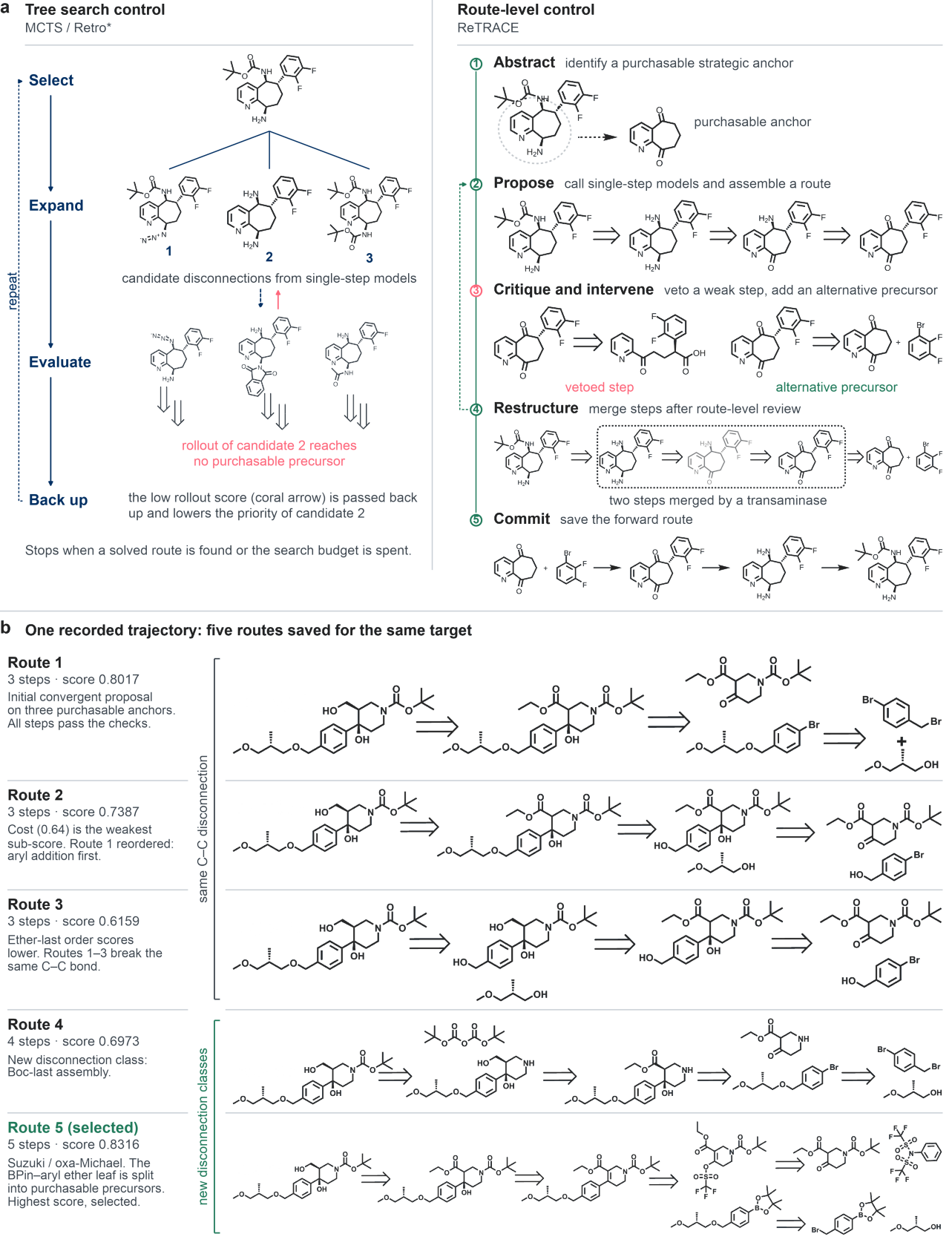}
\caption{\textbf{Conventional tree-search control and ReTRACE route-level control, with one real decision trajectory.} \textbf{a}, Conventional MCTS or Retro* controllers repeatedly select, expand, evaluate and backpropagate within a fixed candidate space. ReTRACE instead runs a agent loop that diagnoses weak steps, introduces alternative precursors and restructures a route before saving it. \textbf{b}, Five routes saved for one representative target, with their total scores. Routes 2 and 3 only reorder the steps of Route 1, Route 4 uses a Boc-last assembly, and Route 5 changes the disconnection class and reaches the highest score (0.8316) after its unpurchasable BPin--aryl ether leaf is disconnected further.}
\label{fig:3}
\end{figure}

Figure~\ref{fig:3}a contrasts two kinds of search control. Conventional MCTS or Retro* controllers expand a search tree within the candidate space defined by single-step models and update node priorities from value estimates or rollout outcomes. ReTRACE keeps the same specialist single-step tools but places branch selection in route-level context: the LLM can compare different reaction strategies, diagnose the current route from purchasability and multi-criterion scores, and backtrack when evidence is insufficient or a local step is unreliable. The difference is a change in how search is controlled, not a substitute for single-step model capability. Figure~\ref{fig:3}b shows a real execution trajectory. The LLM first identified three purchasable strategic anchors, methoxypropanol, 4-bromobenzyl bromide and a \ensuremath{\beta}-keto ester, and assembled convergent Route 1 from them. The two single-step engine classes then gave support scores of 0.96, 1.00 and 0.976 for the Williamson etherification, Grignard addition and ester reduction, respectively, and Route 1 received a total score of 0.8017. The route passed structural and single-step checks, but the route evaluator identified cost as its main weakness. The LLM therefore kept Route 1 while testing whether changing the reaction order would improve the route.

Routes 2 and 3, obtained only by reordering steps, scored 0.7387 and 0.6159, indicating that the problem lay not in the step order but in the shared C--C disconnection strategy. The LLM then switched to a different disconnection class: Route 4 used a Boc-last assembly and scored 0.6973, and Route 5 switched to a Suzuki/oxa-Michael route. When the BPin--aryl ether leaf in Route 5 turned out not to be purchasable, the LLM disconnected it further into 4-BPin-benzyl bromide and methoxypropanol and, after re-checking, obtained the highest-scoring route with a total score of 0.8316 and a feasibility score of 0.90. The case shows that route-level feedback can trigger a directed search that moves from local reordering to a change of disconnection strategy.

\subsection{Tool-group ablation}\label{tool-group-ablation}

We further examined three functionally distinct tool groups: single-step prediction supplies data-driven disconnection candidates, the route scorer feeds back the feasibility, cost and greenness of the full route, and molecular-perception tools provide deterministic structural operations and molecular properties. Each ablated system was run once on Retro*-190 with Claude Opus 4.8 and compared, target by target, with the full system (Table~\ref{tab:performance}). Removing single-step prediction caused the largest and most consistent decrease. This arm was evaluated on 189 targets because one run transcript was misattributed and excluded. On these 189 targets the full system reached Audit@1 of 70.4\% and Audit@10 of 87.8\%, slightly above the 70.0\% and 87.4\% it reached on all 190 targets (Table~\ref{tab:performance}). Without single-step prediction, Audit@1 fell to 42.3\% and Audit@10 to 77.8\% (differences \ensuremath{-}28.0 and \ensuremath{-}10.1 percentage points, paired bootstrap 95\% CI \ensuremath{-}37.0 to \ensuremath{-}19.0 and \ensuremath{-}16.9 to \ensuremath{-}3.2, exact McNemar \emph{p} = 2.9 \ensuremath{\times} 10\textsuperscript{\ensuremath{-}8} and 0.005), and the decrease was significant at all four values of k. This result is consistent with the candidate complementarity in Figure~\ref{fig:2}. Without data-driven single-step suggestions, the coordinator struggles to maintain candidate coverage through general chemical reasoning alone. This supports using the LLM to weigh and combine the outputs of specialized models rather than replacing them with LLM generation.

Removing route scoring lowered Audit@1 by 17.9 percentage points (95\% CI \ensuremath{-}26.8 to \ensuremath{-}8.9, \emph{p} = 2.7 \ensuremath{\times} 10\textsuperscript{\ensuremath{-}4}), but the difference was \ensuremath{-}2.1 points at k = 3 and k = 5 and \ensuremath{-}0.5 points at k = 10 (95\% CI \ensuremath{-}6.3 to 5.3), and no interval at k \ensuremath{\geq} 3 excluded zero. Without route-score feedback, the system found an acceptable route within its top ten candidates about as often but placed it first less often, which is consistent with the role of the scorer in ranking routes. Removing molecular-perception tools lowered Audit@k by 3.2 to 9.5 points, and no difference was significant (\emph{p} \ensuremath{\geq} 0.057). This does not show that molecular perception is unnecessary. With 190 targets, the comparison has limited power for differences of a few percentage points.

\subsection{Prospective wet-lab validation by synthetic chemists}\label{prospective-wet-lab-validation-by-synthetic-chemists}

A computational audit can exclude detectable structural and mechanistic contradictions, but it cannot replace optimization of experimental conditions, nor can it show that a route can be executed as proposed. The prospective validation therefore tests whether route proposals generated by ReTRACE can be translated into executable procedures after review and any necessary revision by synthetic chemists.

\begin{figure}[p]
\centering
\includegraphics[width=\textwidth]{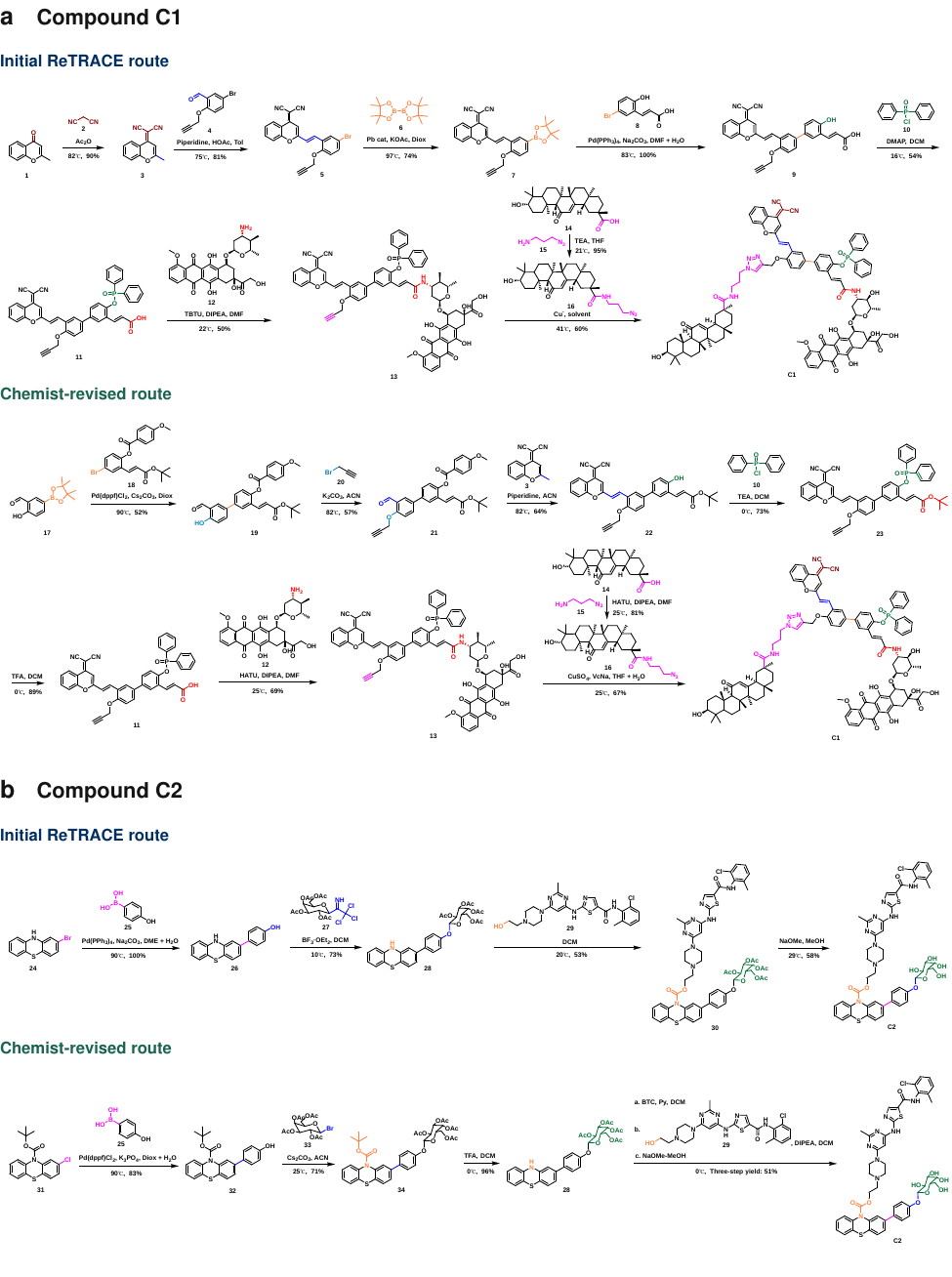}
\caption{\textbf{Prospective wet-lab validation of ReTRACE-assisted route design.} \textbf{a}, Compound C1. \textbf{b}, Compound C2. Each panel compares the initial ReTRACE route with the chemist-revised experimental route. C1 required a revised protection strategy and an earlier Suzuki coupling, whereas C2 retained the core sequence with localized substrate and condition changes.}
\label{fig:4}
\end{figure}

Figure~\ref{fig:4}a shows that, for target compound C1, ReTRACE provided the core retrosynthetic framework and an initial plan that was chemically reasonable overall. Chemist review identified two limitations of the predicted route. First, in the phosphorylation step, the free carboxyl group can compete with the phenolic hydroxyl group in side reactions, lowering selectivity. Second, the Suzuki coupling was placed late in the synthesis, where the sterically hindered substrate would lower the coupling yield and the intramolecular phosphate ester bond would risk cleavage.

To address these issues, the chemists modified the predicted route in two respects, protecting-group strategy and reaction order. To avoid competing side reactions of the carboxyl group during phosphorylation, compound 18, in which the carboxyl group is protected as a tert-butyl ester, was used as the starting substrate. The synthetic order was also changed so that the Suzuki coupling became the first step: cross-coupling of compounds 17 and 18 gave intermediate 19 in 52\% yield, avoiding the efficiency loss of coupling a hindered substrate. Compound 19 then underwent K\textsubscript{2}CO\textsubscript{3}-mediated nucleophilic substitution with 3-bromopropyne (compound 20) to give intermediate 21 in 57\% yield. Following the reaction design proposed by ReTRACE, compound 21 was condensed with compound 3 under piperidine catalysis with concomitant removal of the methoxybenzoate protecting group, giving intermediate 22 in 64\% yield. Intermediate 22 was phosphorylated with phosphoryl chloride 10 in the presence of triethylamine to give intermediate 23 (73\%), and removal of the tert-butyl group with trifluoroacetic acid gave the key intermediate 11 in 89\% yield. The final amide coupling and CuAAC click reaction retained the reaction types of the ReTRACE route, with the coupling reagent changed from TBTU to HATU: intermediate 11 was coupled with doxorubicin using HATU/DIPEA to give intermediate 13 (69\%), and intermediate 13 was then clicked with the separately prepared intermediate 16 under copper sulfate/sodium ascorbate catalysis to give C1 in 67\% yield. The main linear sequence therefore comprises seven isolated transformations; compound 16 was prepared in a separate branch (81\%).

As shown in Figure~\ref{fig:4}b, for target compound C2 the reaction logic, core reaction types and overall synthetic order proposed by ReTRACE were largely retained in the route finally used, with modifications mainly to substrate choice and experimental details.

The first step in both routes builds the biaryl scaffold by Suzuki cross-coupling with 4-hydroxyphenylboronic acid (compound 25), giving intermediate 32 in 83\% yield; in the experiment, the starting material was changed to the Boc-protected amine 31 to avoid amine side reactions during glycosylation and to improve availability. The reaction type and transformation logic of the second step, glycosylation, were also retained; on the basis of practical experience, the chemists used compound 33 as the glycosyl donor and grafted it by caesium carbonate-mediated nucleophilic substitution, giving intermediate 34 in 71\% yield. Removal of the tert-butyl-based protecting group with trifluoroacetic acid (96\%), followed by a three-step sequence of triphosgene-mediated coupling with dasatinib and removal of the acetyl protecting groups with sodium methoxide, gave C2 (51\% over three steps). The core bond-forming logic of the experimental route thus followed the model proposal, with the changes listed above.

In both cases, ReTRACE rapidly produced a workable route skeleton that guided synthesis design, and the targets were obtained after chemistry-informed revisions of varying extent. These two expert-revised demonstrations indicate practical utility; they are not an estimate of the rate of autonomous synthesis success.

\subsection{Failure modes and evidence boundaries}\label{failure-modes-and-evidence-boundaries}

The frozen two-stage audit reveals important limits behind the route-level metrics. Of the 1,250 candidate routes returned by GPT-5.6 Sol standard on Retro*-190, 753 (60.2\%) passed the audit and 497 (39.8\%; 95\% target-bootstrap CI 33.8--45.8\%) failed.

\begin{figure}[!t]
\centering
\includegraphics[width=\textwidth]{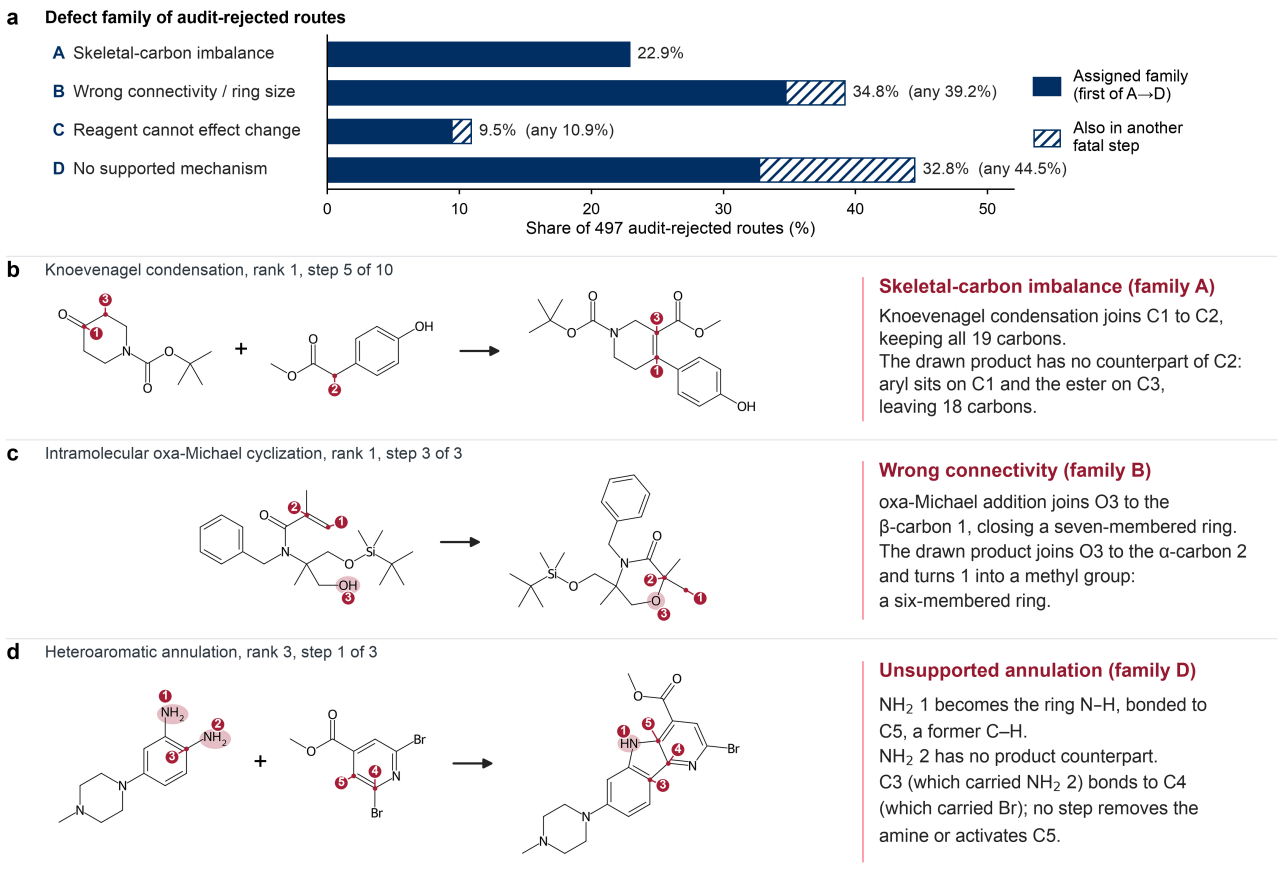}
\caption{\textbf{Chemical failure modes identified by the route-level adversarial audit.} \textbf{a}, Operational grouping of the 497 audit-failed routes of GPT-5.6 Sol (standard) on Retro*-190. Each route is assigned the first family A--D that occurs among its fatal steps (filled bars); hatched extensions add routes that also contain that family but were assigned an earlier one. \textbf{b}, a Knoevenagel condensation should keep all 19 reactant carbons; the drawn product has no counterpart of carbon 2 and contains 18. \textbf{c}, an intramolecular oxa-Michael addition should join alcohol oxygen 3 to \ensuremath{\beta}-carbon 1 and close a seven-membered ring; the drawn product joins it to \ensuremath{\alpha}-carbon 2 and turns carbon 1 into a methyl group, giving a six-membered ring. \textbf{d}, a one-step construction of a \ensuremath{\delta}-carboline (pyrido{[}3,2-\emph{b}{]}indole) from a phenylenediamine and a dibromopyridine ester forms new C--N and C--C bonds while the nitrogen count falls from five to four, with no nitrogen-containing leaving group.}
\label{fig:5}
\end{figure}

Failed routes were not dominated by formatting or parsing problems; they concentrated on chemical contradictions within the route. The audit records a defect code for every fatal step, and a route can contain several fatal steps of different kinds, so the original codes overlap at the route level. We therefore grouped the codes into four operational families and assigned each route to the first family it contains in a fixed order, from the most directly checkable evidence to the most interpretive: (A) skeletal carbons appear or vanish with no reagent to supply or remove them, (B) the drawn product has the wrong connectivity, regiochemistry or ring size, (C) the stated or implied reagent cannot effect the drawn change, and (D) no known mechanism links the drawn reactants to the drawn product. Of the 497 failed routes, 22.9\% (95\% CI 15.4--30.9\%) were assigned to A, 34.8\% (26.7--43.3\%) to B, 9.5\% (5.7--14.1\%) to C and 32.8\% (25.4--40.6\%) to D. Eighty-one routes contain fatal steps from more than one family. Figure~\ref{fig:5} places this distribution alongside three real audit cases, one each from families A, B and D. Such errors can be confined to one step or propagate to later steps, so that a structurally closed route can still fail chemically.

\subsection{Preference steering, human-in-the-loop continuation, chemo-enzymatic extension and post-training}\label{preference-steering-human-in-the-loop-continuation-chemo-enzymatic-extension-and-post-training}

Beyond core planning, we examined three extensions in case studies: changing preference rankings within a fixed candidate pool by adjusting route weights; accepting chemist feedback and continuing planning from persisted route state; and calling an explicit enzymatic single-step tool so that enzymatic disconnections with provenance are retained in complete candidates.\cite{ref32} These results are given in Appendix~\ref{app:cases}.

The ReTRACE harness provides evaluation signals during inference, and the same signals could in principle serve as rewards or targets for training, so that experience gained during planning is distilled into model parameters. We therefore saved the states, actions, tool returns and termination outcomes recorded by the harness as trajectories for post-training and tried three methods: supervised fine-tuning (SFT), GRPO\cite{ref33} and on-policy self-distillation (OPSD)\cite{ref34}. In OPSD the model generates its own continuations from harness decision points, and the same base model, given a verified reference route for the training target, acts as teacher. SFT and GRPO were trained from Qwen3.5-9B, and OPSD was trained from Qwen3.8-27B with LoRA on 16 samples from four training targets. Each arm was compared only with its own base model on the same 40 held-out Retro*-190 targets, and the arms are not ranked against one another because their base models differ. Neither SFT nor GRPO exceeded its historical base. Under our evaluation metric, OPSD produced 9 of 40 rank-1 routes that passed the audit against 16 of 40 for its base (exact McNemar \emph{p} = 0.039, paired difference \ensuremath{-}17.5 percentage points, 95\% CI \ensuremath{-}30 to \ensuremath{-}5). We leave harness-informed post-training as a direction for future work. Training configurations and results are given in Appendix~\ref{app:posttrain}.

\section{Methods}\label{methods}

\subsection{Task formulation and system overview}\label{task-formulation-and-system-overview}

We formulate multistep retrosynthetic planning as a tool-augmented sequential decision problem. Given the SMILES of a target molecule, a library of commercially purchasable starting materials and a set of single-step expansion tools, an LLM coordinator recursively selects disconnections until all leaves of the synthesis tree fall within the purchasable library. The search state is a triple consisting of the current synthesis tree, the external route memory and the remaining budget; each action is one tool call; and a planning task terminates in one of three cases: all leaves are purchasable, the budget is exhausted, or the route memory has accumulated the maximum number of structurally distinct candidate routes.

Figure~\ref{fig:1}a gives the overall architecture of ReTRACE. The chemistry tool layer is organized into four classes according to the type of evidence they provide to the LLM: molecular operations (RDKit-based structural validity checks and template application), single-step retrosynthesis (template-based and template-free engines), starting-material purchasability lookup and three-dimensional route scoring (feasibility, cost and greenness). The coordinator is an LLM that works in rounds. In each round it selects the molecule to disconnect and writes a chemical rationale, calls the selected tools, and then inspects the returned precursors and scores to decide whether to continue disconnecting, switch branches or backtrack. The route memory sits outside the loop; it stores constructed partial routes, previous disconnection decisions and abandoned branches, and is injected as prior context into the next round of reasoning, so that maintaining the search state does not depend on the model's finite context window.

An LLM is a probabilistic language model, and its outputs carry no inherent structural reliability. Every interaction between the model and the tools, and between the model and the memory, therefore passes through deterministic guarantee gates, which are enforced by system code. The action gate rejects illegal operations and unparsable SMILES before a request reaches a tool. The result gate performs preliminary screening and fingerprint-based ensemble de-duplication before candidate precursors are returned to the LLM, and the admission gate ensures that only fully scored routes can be written to route memory.

For every accepted or rejected disconnection, ReTRACE records (i) which tool class proposed the candidate precursors and what they were; (ii) the route-level evaluation feedback actually visible to the LLM when it made the decision; and (iii) the reason the LLM stated for accepting, modifying or abandoning the branch. These records ensure decision provenance and replayability. Figure~\ref{fig:1}b,c excerpts two decision instances from real search trajectories, adjudication of conflicting single-step suggestions and diagnosis and backtracking after an unfavourable score, which show these three kinds of records at single decision points.

\subsection{Operation tools: molecular operations, single-step expansion and purchasability lookup}\label{operation-tools-molecular-operations-single-step-expansion-and-purchasability-lookup}

A basic difficulty in automated retrosynthesis is that no single single-step model covers the entire space of reaction types. Template-based methods are precise for common transformations within the training distribution but fail on novel scaffolds, whereas template-free neural methods generalize better but sometimes produce mechanistically problematic products. The tool layer of ReTRACE is therefore deliberately heterogeneous and complementary rather than built around a single best engine. This complementarity assumption is tested quantitatively on the 50,000 reactions of USPTO-50K (Figure~\ref{fig:2}).

The operation tools fall into three groups, organized by the type of evidence they provide to the LLM.

\emph{Deterministic molecular operations.} A set of RDKit-based deterministic tools performs SMILES canonicalization, structural validity checks, disconnection analysis and template application.\cite{ref35} These tools provide structure-level chemical guardrails, so that the validity of SMILES does not depend on the internal representations of the language model.

\emph{Single-step expansion engines.} Single-step expansion is offered through two complementary entries. The ensemble entry queries five heterogeneous engines in parallel under a unified interface: GraphFP, a fingerprint-based graph neural model trained on USPTO-Full; RXNGraphormer, a sequence/graph model trained on USPTO-Full; ChemBART, a pretrained Transformer (BART architecture) pretrained with a reaction-level mask-infilling objective on USPTO-full and USPTO-MIT; a template-relevance model based on the Reaxys template library; and the ring-opening template specialist Ringbreaker. Each ensemble engine emits up to 50 candidate precursors. After cross-engine canonicalization, de-duplication and diversity filtering, about 15 diverse candidates are returned to the LLM, each annotated with a cross-engine agreement (support) field. The LLM is instructed to query both entries in parallel for every non-leaf molecule in the same round rather than selecting a disconnection from its chemical intuition alone.

\emph{Purchasability lookup.} Route termination is determined by two canonicalized stock libraries consistent with ChemEnzyRetroPlanner: BioNav (103,530 canonicalized molecules) and ZINC (10,320,667 canonicalized molecules).

The LLM can combine these tools through programmatic tool calling in the harness. Unlike conventional tool calling in agents, the LLM can compose multiple tools more flexibly in code, for example calling several single-step retrosynthesis tools at once and keeping only the purchasable part of their results. Intermediate tool states (here, the full result returned by each single-step tool) then do not need to enter the LLM context window, which saves tokens and shortens response time.

\subsection{Scoring tools: multi-criterion route evaluation and closed-loop feedback}\label{scoring-tools-multi-criterion-route-evaluation-and-closed-loop-feedback}

A single feasibility score cannot distinguish a route that is chemically feasible but long and dependent on regulated reagents from a route that is short and uses only green reagents, and scoring after the fact cannot influence the search itself. The route evaluator of ReTRACE is therefore multi-criterion and runs inside the search loop: it is used both to evaluate routes and to guide search.

The route evaluator outputs a weighted total score composed of three dimensions: feasibility, cost and greenness. Users can adjust the three weights to suit different project goals; all main experiments in this work use the default 0.70/0.15/0.15, which prioritizes feasibility while giving equal weight to cost and greenness. The weights change only candidate comparison during search and the final ranking. They do not change the independent audit's pass/fail verdict on a route.

Feasibility is scored for each step by a pretrained discriminative reaction classifier.\cite{ref18} Feasibility is not a simple average of the step scores: the mean is shifted towards the worst step of the route to penalize routes containing an individual severe defect. It is complemented by a penalty for fabricated carbon skeletons (applied when the product contains more carbons than the reactants combined) and by flags for stereochemical anomalies. In practice we found that the pretrained reaction classifier can assign very low scores to steps that lie outside its training distribution but are chemically sound (for example, near-zero feasibility for a textbook Boc deprotection or for displacement of a primary sulfonate by a secondary amine). The system therefore includes an LLM review safety net that is triggered only when the feasibility score of a step or a whole route falls below a threshold, can only raise sub-threshold scores and is bounded by a cap (threshold and cap values in Appendix~\ref{app:review}). The review keeps the original score by default and outputs a corrected score only when it judges the classifier to have produced a clear false negative. In a paired three-arm ablation, score changes were confined to steps with classifier confidence below the trigger threshold and affected 23.1\% of all steps on JMC2025 and 10.7\% on an FDA set. Among rank-1 routes, audit reportability was similar with and without a review-lifted step (70.3\% versus 74.5\%; Fisher two-sided \emph{p} = 0.687); this selected-route check does not establish that every upward correction recovers a genuine false negative.

Cost combines three signals: a step-efficiency term defined to decrease monotonically with the number of synthetic steps; leaf purchasability and price signals; and the reduction in synthetic complexity along the route, measured as the drop in SCScore of Coley et al.\cite{ref20} from product to precursors.

Greenness combines three terms: atom economy (Trost\cite{ref37}), solvent classification based on the CHEM21 solvent selection guide\cite{ref38} and a SMARTS scan for hazardous reagents.

\emph{Closed-loop feedback.} Candidate ranking during search and final route scoring share the same weights; the former is used when choosing disconnections and the latter when evaluating routes. As a direct consequence of this loop, the evaluator becomes search feedback rather than a post hoc score: the LLM backtracks in a targeted way from the lowest-scoring dimension, rather than resampling blindly, and adjusts its planning strategy in real time from the scorer's feedback.

\subsection{LLM coordinator, route memory and context compaction}\label{llm-coordinator-route-memory-and-context-compaction}

The context window of a language model is finite, and the model keeps no persistent state between calls; a multistep search spanning dozens of rounds of tool interaction cannot keep the synthesis tree, the record of failed branches and the ranking of candidate routes in the model's internal representation. ReTRACE therefore externalizes the entire search state as an explicit route memory and runs an agent loop on top of it.

\emph{Route memory.} The route memory offers the LLM three operations: submitting a candidate route (de-duplicated by route fingerprint, which rejects trivial variants of saved routes), retrieving saved candidates by total score, and recording the reason and lowest sub-score of a branch before it is abandoned. Every saved record carries the route label, total score, reaction steps, dimension sub-scores, starting materials and free-text notes; the system also records an internal provenance tag for each route that is not visible to the LLM and is used only for post hoc analysis. The route memory keeps evaluated branches and exposes them to the LLM in later decisions.

\emph{Termination and budget.} Search terminates when every leaf of the current tree is purchasable, when the single-step expansion query budget is exhausted, or when the number of saved, structurally distinct routes reaches the preset capacity. The maximum search depth is 12. Only queries to the two single-step expansion entries consume budget, and a pair of queries issued in parallel in the same round counts as one call. The main results use the heterogeneous presets frozen in the run-level configurations: GPT-5.6 Sol fast, standard and thorough use medium/high/xhigh reasoning efforts, 50/100/200 queries, route capacities of 10/10/20 and wall-clock limits of 4/6/10 h. Each model keeps the same preset on Retro*-190, JMC2025 and FDA30.

\emph{Context compaction and trajectory logging.} Because the finite context window of an LLM cannot hold tasks that may run for hours, we introduced a context-compaction mechanism. Compaction trims only the LLM's active context, namely the operating instructions, the target molecule and a snapshot of the route memory, to free the window for new exploration. The complete search trajectory, including raw tool outputs, evaluator feedback and backtracking rationales, is persisted to disk together with timestamped markers of compaction events. Compaction deletes no audit evidence: it moves large amounts of information that may no longer be relevant out of the LLM context window and replaces it with links to the corresponding disk locations, which the LLM can retrieve at any time through a memory tool.

\subsection{Benchmarks, baselines and two-stage audit protocol}\label{benchmarks-baselines-and-two-stage-audit-protocol}

A failure mode specific to LLM agents is the generation of routes that look plausible and conserve atoms yet are mechanistically impossible, errors that conventional top-k exact matching cannot capture. The evaluation protocol must therefore be independent of the system's internal evaluator. We first describe this structural separation and then the benchmark construction, the comparison systems and the two-stage audit.

\emph{Benchmark datasets.} We use three complementary benchmarks. Retro*-190 (\emph{n} = 190) is a set of difficult target molecules selected from the Retro* test set and serves as the standard benchmark. JMC2025 (\emph{n} = 141) is drawn from medicinal-chemistry literature published in the \emph{Journal of Medicinal Chemistry} in 2025. FDA30 (\emph{n} = 30) consists of small-molecule drugs approved by the FDA between 2020 and 2024.

\emph{Baselines.} External comparisons include classical Retro* search, ReTRACE coordinated by the open-weight LLM Qwen3.6-35B-A3B, and direct generation by Codex, one of the strongest general-purpose agents. All comparisons use the same target sets, frozen stock-termination criterion, route canonicalization and two-stage audit, but keep their native controllers and budgets: Retro* returns at most one official search route per target; Codex performs one isolated call per target with a 90 min wall-clock limit.

\emph{Two-stage audit protocol.} Route-level evaluation uses a two-stage audit applied with the same model and the same prompt templates to all backbones, baselines and datasets.

The audit first applies deterministic RDKit gating: products and precursors are parsed and canonicalized step by step, the check verifies that the target is produced by some step of the route, and changes in skeletal carbon count and ring size are recorded. Routes that cannot be parsed or do not produce the target are recorded as audit failures directly. Carbon-count or ring changes are passed to the subsequent review as structured hints. The remaining routes are read step by step, in independent calls, by a fixed single LLM judge (Claude Opus 5, temperature = 0.2, max\_tokens = 32,000), which judges each step as passed or failed and gives a defect rationale. A route with any failed step provisionally fails, and all other routes pass. Only provisionally failed routes trigger a second, adversarial defence: in a fresh context and with the opposite objective, the same judge reads the route together with its failed steps and their rationales, and argues as strongly as possible whether these steps could be realized by real chemistry. It does not read the rest of the first-stage output. If the defence fails, the route fails the audit. If it succeeds, the route passes.

Our main metric, Audit@k, is defined as follows. For a target molecule, candidate routes are sorted from high to low by evaluator score; the target passes at depth k if at least one of the top k routes passes the two-stage audit. If none of the top k routes passes, the target is recorded as a failure. The denominator is always the full number of targets in the dataset, and targets whose runs failed or returned no route also count as failures. Sampling uncertainty is estimated on the per-target binary success vector of the full dataset by percentile bootstrap resampling, giving 95\% confidence intervals (B = 10,000; seed = 42). The complete interval matrix is given in Appendix~\ref{app:ci}.

\emph{Chemist review.} To test the agreement of the automatic audit with human chemical judgement, a researcher with a background in synthetic chemistry reviewed 59 candidate routes from the first six Retro*-190 targets, including all 24 rank-1 routes under four coordinator conditions and selected lower-ranked candidates, and judged 274 reaction steps. The reviewer could see the automatic verdict and rationale, so the review was targeted and non-blind. The primary analysis uses the same binary audit-pass/audit-fail endpoint as Audit@k and reports observed agreement and Cohen's \ensuremath{\kappa}.

\subsection{Wet-lab validation}\label{wet-lab-validation}

\emph{Prospective design.} To test experimental executability beyond the computational audit, we carried out prospective wet-lab validation on two target molecules, C1 and C2, that were not part of any benchmark set (\emph{n} = 2). These two were all the targets that entered experimental validation in this study; no other target was excluded after route generation because of prediction quality or experimental outcome. For each target, ReTRACE generated candidate routes before wet-lab work began.

\emph{Route review and experimental translation.} Synthetic chemists then reviewed the candidate routes provided by ReTRACE for functional-group compatibility, protecting-group strategy, substrate availability, steric effects and reaction order, and translated them into executable procedures. They were allowed, on grounds of chemical rationale and practical operation, to substitute substrates with appropriate protecting groups or better availability, to change the reaction order and to optimize reagents and operating conditions; every change was recorded, with its chemical rationale, relative to the frozen route record. The validation therefore tests whether the core retrosynthetic framework and key bond-forming logic proposed by ReTRACE can lead to the target product after necessary chemist adaptation, not whether the experiment follows the model's suggestion verbatim.

\emph{Execution, characterization and evaluation.} Syntheses were carried out step by step according to the reviewed procedures; reaction progress was monitored by thin-layer chromatography, products were isolated by conventional work-up and silica-gel column chromatography, and the isolated yield of each step was recorded. Key intermediates were characterized by \textsuperscript{1}H NMR, and the structures of the final products C1 and C2 were confirmed by \textsuperscript{1}H NMR, \textsuperscript{13}C NMR and high-resolution mass spectrometry (HRMS). The evaluation record includes whether the target product was isolated and structurally confirmed, the extent to which the key disconnection and bond-forming logic proposed by the model was retained in the experimental route, the types of chemist revision and the yield of each step. Experimental validation was recorded as successful only when the target product was isolated and its structure was confirmed by these analyses. Complete experimental procedures, compound characterization data and spectra are given in Appendix~\ref{app:wetlab}.

\clearpage

\appendix
\renewcommand{\thesection}{\Alph{section}}
\setcounter{figure}{0}\setcounter{table}{0}
\renewcommand{\thefigure}{A\arabic{figure}}
\renewcommand{\thetable}{A\arabic{table}}
\renewcommand{\theHfigure}{appendix.\arabic{figure}}
\renewcommand{\theHtable}{appendix.\arabic{table}}
\makeatletter\setlength{\@fptop}{0pt}\makeatother
\begin{center}{\Large\bfseries Appendix}\end{center}
\vspace{2pt}

\section{Preference steering, human feedback and enzymatic tools}\label{app:cases}

These three case studies probe interfaces beyond the core benchmark. They are case-based and are not estimates of average improvement across targets.

\emph{Preference steering.} On a frozen GPT-5.6 Sol standard candidate pool, a cost-priority reweighting changed the rank-1 route of target 100 from the 8-step default route to the original rank-8 six-step route, which removes a separate aryl-iodide-to-boronate sequence; the cost sub-score increased by 0.0347 and greenness by 0.0101, while feasibility decreased by 0.0285. A green-priority reweighting on target 101 selected an alternative without DIAD/PPh\textsubscript{3}; greenness increased by 0.0841, while feasibility and cost decreased by 0.0543 and 0.0557. Both selected alternatives had passed the audit.

\emph{Human-in-the-loop continuation.} A same-state fork on target ENPP1 A25 gave a no-feedback arm and a chemistry-feedback arm 12 further iterations each. Both arms kept the same rank-1 route and score (0.8042); the feedback arm added one lower-scoring route. Accepting feedback therefore does not guarantee a change of selection. In the prospective C1/C2 work, chemists supplied written route-level feedback that was incorporated into the experimental routes (Figure~\ref{fig:4}).

\emph{Chemo-enzymatic extension.} An explicit enzymatic single-step tool queried a library of 491,622 enzymatic templates and carried source, reaction kind and retrieval rank into the saved-step provenance. For ephedrine, the mixed arm retained a two-step route consistent with the literature topology: acetaldehyde and benzaldehyde are joined by a RetroBioCat TDL carboligation template to give chiral phenylacetylcarbinol, followed by chemical reductive amination with methylamine. The enzymatic step was the rank-1 retrieval for that intermediate, all three leaves were in stock, and the route scored 0.7306 (rank 6 of 10). Under the same GPT-5.6 Sol standard coordinator and at most 18 single-step calls, the organic-only control returned only organic templates among 70 inspected suggestions and did not recover this sequence. Other cases were mixed: rasagiline retained a sourced imine-formation/reduction route but not the literature transaminase--N-propargylation order; for a fluoropyridinyl tryptoline a sourced Pictet--Spengler step was retrieved without the halogenase--Suzuki--cyclization strategy; and for \emph{o}-chloromandelic acid the aldehyde--cyanohydrin--acid topology and a RetroBioCat hydrocyanation source were recovered, but the chemical hydrolysis of the reference route was treated as enzymatic nitrile hydrolysis.

\clearpage
\section{Post-training on harness trajectories}\label{app:posttrain}

Table~\ref{tab:A-posttrain} summarizes the three arms. All arms were evaluated on the same 40 held-out Retro*-190 targets with the frozen two-stage audit (one planning run, one judge sample, original rank-1 route only). Each arm is compared only with its own base model.

For OPSD, one target returned no candidate. The paired differences against its base were $-17.5$ percentage points (exact McNemar \emph{p} = 0.039; paired target-bootstrap 95\% CI $-30$ to $-5$). Under a stricter criterion that also rejects routes the audit marks as suspect (which Audit@1 counts as passed), OPSD reached 2/40 against 4/40 for its base (\emph{p} = 0.63).

\noindent\begin{minipage}{\textwidth}
\begin{center}
\captionsetup{hypcap=false}
\captionof{table}{Post-training configurations and held-out results (N = 40). Audit@1 counts targets whose rank-1 route passed the audit.}
\label{tab:A-posttrain}
\small
\setlength{\tabcolsep}{4pt}
\renewcommand{\arraystretch}{1.2}
\begin{tabular}{@{}>{\raggedright\arraybackslash}p{12mm}>{\raggedright\arraybackslash}p{21mm}>{\raggedright\arraybackslash}p{78mm}>{\centering\arraybackslash}p{19mm}>{\centering\arraybackslash}p{19mm}@{}}
\toprule
\textbf{Arm} & \textbf{Base model} & \textbf{Training data and configuration} & \textbf{Audit@1 (arm)} & \textbf{Audit@1 (base)} \\
\midrule
SFT & Qwen3.5-9B & 1,211 harness-trajectory windows of up to 65k tokens; LoRA; 8 $\times$ A800; 1 epoch; 152 optimizer steps & 24/40 & 27/40 \\
GRPO & Qwen3.5-9B & Offline, on frozen trajectories with message-level relative advantages; 200 optimizer steps; longest sequence 42,027 tokens & 26/40 & 27/40 \\
OPSD & Qwen3.8-27B & Official TRL SDFT trainer; LoRA rank 32; FSDP2 on 2 GPUs; 16 trajectories from 4 training targets; 1 epoch; 8 optimizer steps; learning rate $2\times10^{-5}$; seed 42 & 9/40 & 16/40 \\
\bottomrule
\end{tabular}
\end{center}
\end{minipage}

\section{LLM review safety net: trigger and caps}\label{app:review}

\begin{itemize}\setlength{\itemsep}{1pt}
\item \emph{Trigger.} Review is invoked only when the feasibility score of a step, or the aggregated feasibility of a route, is below 0.4. Scores at or above 0.4 are never reviewed.
\item \emph{Uphold by default.} The review first decides whether the low classifier score is justified. If it is, the original score is kept unchanged; a corrected score is emitted only for a clear false negative.
\item \emph{Caps.} A corrected score is capped at 0.85 for a step and at 0.7 for a route, and the final score is the maximum of the original and the capped corrected score, so review can only raise a score and cannot lift a step to a perfect score.
\item \emph{Fail-closed.} If the review output cannot be parsed, times out or the service is unavailable, the original classifier score is kept.
\end{itemize}

\clearpage
\section{Audit@\texorpdfstring{$k$}{k} confidence intervals}\label{app:ci}

\par\vspace{4pt}
\noindent\begin{minipage}{\textwidth}
\begin{center}
\captionsetup{hypcap=false}
\captionof{table}{Audit@$k$ (\%) with 95\% confidence intervals for all controllers, baselines and Retro*-190 ablations in Table~\ref{tab:performance}. Intervals are percentile bootstrap over targets (B = 10,000; seed = 42). Retro* and Codex return at most one route, so their values are constant in $k$.}
\label{tab:A-ci}
\small
\setlength{\tabcolsep}{4pt}
\renewcommand{\arraystretch}{1.12}
\begin{tabular}{@{}>{\raggedright\arraybackslash}p{50mm}cccc@{}}
\toprule
\textbf{Controller / baseline} & \textbf{$k=1$} & \textbf{$k=3$} & \textbf{$k=5$} & \textbf{$k=10$} \\
\midrule
\multicolumn{5}{@{}l}{\textbf{Retro*-190 (N = 190)}} \\
Claude Opus 4.8 & 70.0 [63.2, 76.3] & 78.4 [72.1, 84.2] & 83.7 [78.4, 88.9] & 87.4 [82.6, 91.6] \\
GPT-5.6 Sol fast & 64.2 [57.4, 71.1] & 76.8 [70.5, 82.6] & 78.4 [72.6, 84.2] & 80.0 [74.2, 85.8] \\
GPT-5.6 Sol standard & 65.8 [58.9, 72.6] & 75.3 [68.9, 81.1] & 79.5 [73.7, 85.3] & 82.6 [76.8, 87.9] \\
GPT-5.6 Sol thorough & 63.7 [56.8, 70.5] & 72.6 [66.3, 78.9] & 79.5 [73.7, 85.3] & 82.1 [76.3, 87.4] \\
Qwen3.6-35B-A3B & 23.7 [17.9, 30.0] & 33.2 [26.3, 40.0] & 33.7 [26.8, 40.5] & 34.7 [27.9, 41.6] \\
\addlinespace[2pt]
Retro* & 73.7 [67.4, 80.0] & 73.7 [67.4, 80.0] & 73.7 [67.4, 80.0] & 73.7 [67.4, 80.0] \\
Codex & 32.1 [25.8, 38.9] & 32.1 [25.8, 38.9] & 32.1 [25.8, 38.9] & 32.1 [25.8, 38.9] \\
\addlinespace[2pt]
w/o single-step prediction\textsuperscript{a} & 42.3 [35.4, 49.2] & 57.7 [50.3, 64.6] & 68.8 [61.9, 75.7] & 77.8 [72.0, 83.6] \\
w/o route scoring & 52.1 [45.3, 59.5] & 76.3 [70.0, 82.1] & 81.6 [75.8, 86.8] & 86.8 [81.6, 91.6] \\
w/o molecular-perception tools & 60.5 [53.7, 67.4] & 73.2 [66.3, 78.9] & 76.8 [70.5, 82.6] & 84.2 [78.9, 88.9] \\
\addlinespace[5pt]
\multicolumn{5}{@{}l}{\textbf{JMC2025 (N = 141)}} \\
Claude Opus 4.8 & 82.3 [75.9, 87.9] & 87.9 [82.3, 92.9] & 90.1 [85.1, 95.0] & 95.0 [91.5, 98.6] \\
GPT-5.6 Sol fast & 75.2 [68.1, 82.3] & 85.8 [80.1, 91.5] & 88.7 [83.0, 93.6] & 89.4 [84.4, 94.3] \\
GPT-5.6 Sol standard & 75.9 [68.8, 83.0] & 85.8 [80.1, 91.5] & 88.7 [83.0, 93.6] & 92.2 [87.2, 96.5] \\
GPT-5.6 Sol thorough & 75.2 [68.1, 82.3] & 85.1 [78.7, 90.8] & 85.8 [80.1, 91.5] & 88.7 [83.0, 93.6] \\
Qwen3.6-35B-A3B & 24.8 [17.7, 31.9] & 39.7 [31.9, 48.2] & 42.6 [34.8, 50.4] & 43.3 [35.5, 51.8] \\
\addlinespace[2pt]
Retro* & 80.9 [74.5, 87.2] & 80.9 [74.5, 87.2] & 80.9 [74.5, 87.2] & 80.9 [74.5, 87.2] \\
Codex & 40.4 [32.6, 48.2] & 40.4 [32.6, 48.2] & 40.4 [32.6, 48.2] & 40.4 [32.6, 48.2] \\
\addlinespace[5pt]
\multicolumn{5}{@{}l}{\textbf{FDA30 (N = 30)}} \\
Claude Opus 4.8 & 70.0 [53.3, 86.7] & 80.0 [63.3, 93.3] & 83.3 [70.0, 96.7] & 86.7 [73.3, 96.7] \\
GPT-5.6 Sol fast & 70.0 [53.3, 86.7] & 80.0 [63.3, 93.3] & 83.3 [70.0, 96.7] & 86.7 [73.3, 96.7] \\
GPT-5.6 Sol standard & 76.7 [60.0, 90.0] & 83.3 [70.0, 96.7] & 86.7 [73.3, 96.7] & 90.0 [76.7, 100.0] \\
GPT-5.6 Sol thorough & 66.7 [50.0, 83.3] & 76.7 [60.0, 90.0] & 76.7 [60.0, 90.0] & 83.3 [70.0, 96.7] \\
Qwen3.6-35B-A3B & 30.0 [13.3, 46.7] & 53.3 [33.3, 70.0] & 60.0 [43.3, 76.7] & 63.3 [46.7, 80.0] \\
\addlinespace[2pt]
Retro* & 53.3 [36.7, 70.0] & 53.3 [36.7, 70.0] & 53.3 [36.7, 70.0] & 53.3 [36.7, 70.0] \\
Codex & 30.0 [13.3, 46.7] & 30.0 [13.3, 46.7] & 30.0 [13.3, 46.7] & 30.0 [13.3, 46.7] \\
\bottomrule
\end{tabular}

\vspace{3pt}
{\footnotesize\raggedright \textsuperscript{a}One target had no adjudicated audit outcome in this arm; values use the 189 adjudicated targets, as in Table~\ref{tab:performance}.\par}
\end{center}
\end{minipage}
\vspace{2pt}

\clearpage
\section{Wet-lab procedures, characterization data and spectra}\label{app:wetlab}

The initial ReTRACE routes and the chemist-modified experimental routes are shown in Figure~\ref{fig:4}; compound numbers below follow that figure. Each step of the experimental route is given with its procedure and characterization data. Yields are isolated yields unless stated as crude. Spectra are shown in Figures~\ref{fig:A-specC1} and \ref{fig:A-specC2}.

\subsection*{Compound C1}

\begin{center}\includegraphics[width=0.49\textwidth]{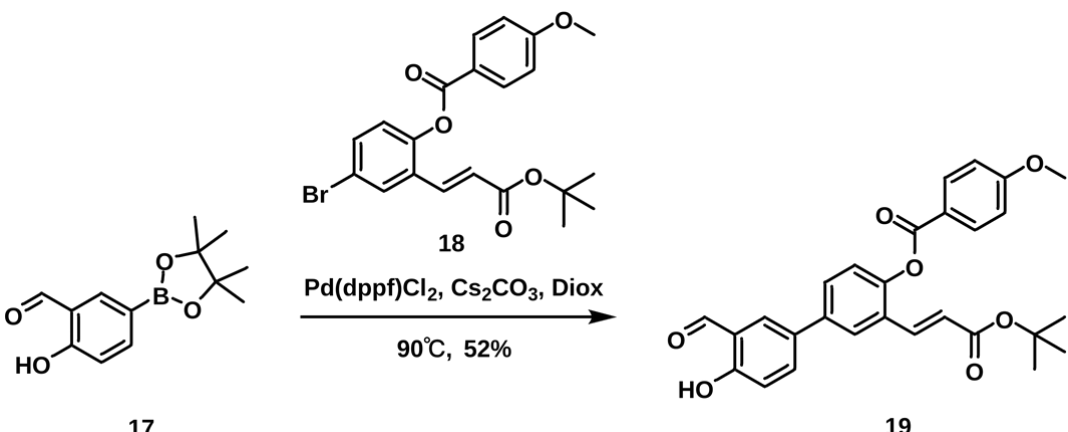}\end{center}\nopagebreak[4]
\vspace{-6pt}\noindent To a side-arm ground-glass Schlenk flask, compound 17 (1.89 g, 7.62 mmol), compound 18 (2.20 g, 5.08 mmol), Pd(dppf)Cl\textsubscript{2} (0.18 g, 0.25 mmol), and Cs\textsubscript{2}CO\textsubscript{3} (3.31 g, 10.15 mmol) were added sequentially. 1,4-Dioxane (Diox, 15 mL) was added as the reaction solvent. The reaction system was evacuated and backfilled with nitrogen three times via a Schlenk manifold, and the mixture was then stirred at 90~\textdegree C for 16 h. Upon completion of the reaction, the mixture was filtered through a Celite pad. The filtrate was concentrated under reduced pressure, and the crude product was purified by silica gel column chromatography to afford compound 19 as a pale yellow solid (1.01 g, 52\% yield). \textsuperscript{1}H NMR (500 MHz, Chloroform-d) $\delta$ 10.03 (s, 1H), 8.25 -- 8.19 (m, 2H), 7.88 -- 7.77 (m, 4H), 7.62 (dd, J = 8.4, 2.3 Hz, 1H), 7.35 (d, J = 8.4 Hz, 1H), 7.16 -- 7.10 (m, 1H), 7.06 -- 7.01 (m, 2H), 6.51 (d, J = 16.0 Hz, 1H), 3.94 (s, 3H), 1.50 (s, 9H).

\begin{center}\includegraphics[width=0.5\textwidth]{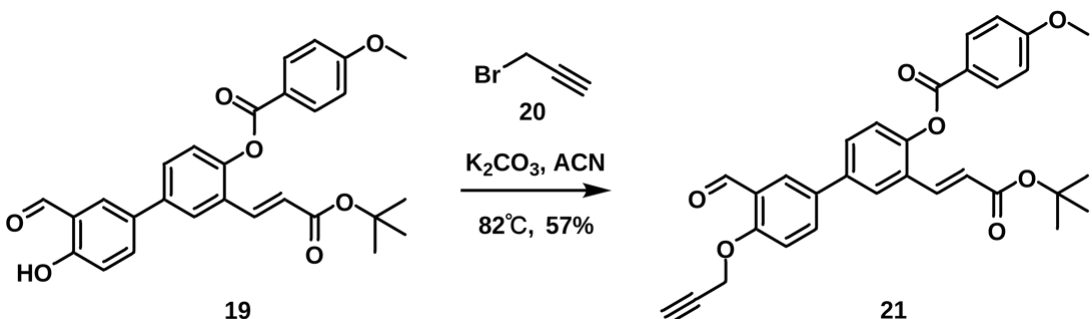}\end{center}\nopagebreak[4]
\vspace{-6pt}\noindent Compound 19 (0.96 g, 2.02 mmol) was dissolved in acetonitrile (ACN, 12 mL), and K\textsubscript{2}CO\textsubscript{3} (0.56 g, 4.05 mmol) was added. After stirring for 30 min, 3-bromopropyne (compound 20, 0.23 mL, 3.03 mmol) was added dropwise. The reaction mixture was heated to 82~\textdegree C and stirred under a nitrogen atmosphere for 6 h. After completion of the reaction was confirmed by thin-layer chromatography (TLC) monitoring, the mixture was filtered through a Celite pad. The filtrate was concentrated under reduced pressure, and the crude product was purified by silica gel column chromatography to afford compound 21 as a pale yellow solid (0.59 g, 57\% yield). \textsuperscript{1}H NMR (500 MHz, Chloroform-d) $\delta$ 10.56 (s, 1H), 8.24 -- 8.20 (m, 2H), 8.13 (d, J = 2.5 Hz, 1H), 7.88 (d, J = 2.3 Hz, 1H), 7.85 -- 7.78 (m, 2H), 7.64 (dd, J = 8.4, 2.3 Hz, 1H), 7.33 (d, J = 8.5 Hz, 1H), 7.25 (d, J = 8.7 Hz, 1H), 7.05 -- 7.01 (m, 2H), 6.51 (d, J = 16.0 Hz, 1H), 4.92 (d, J = 2.4 Hz, 2H), 3.93 (s, 3H), 2.64 (t, J = 2.4 Hz, 1H), 1.50 (d, J = 10.3 Hz, 9H).

\clearpage
\begin{center}\includegraphics[width=0.59\textwidth]{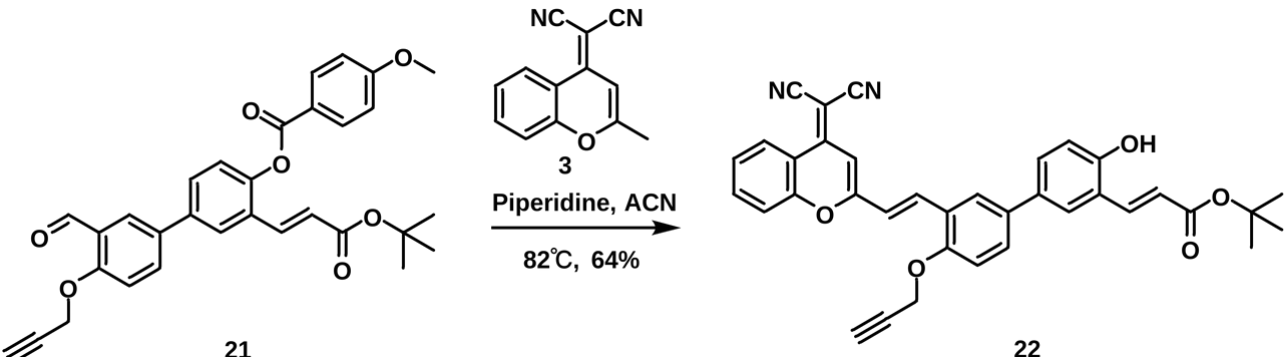}\end{center}\nopagebreak[4]
\vspace{-6pt}\noindent Compound 21 (0.59 g, 1.15 mmol) and compound 3 (0.24 g, 1.15 mmol) were dissolved in ACN (10 mL). After stirring until a homogeneous solution was formed, piperidine (0.17 mL, 1.73 mmol) was added dropwise slowly. The reaction mixture was then heated to 82~\textdegree C and stirred for 4 h under a nitrogen atmosphere. The reaction progress was monitored by TLC. Upon complete consumption of the starting materials, the mixture was concentrated under reduced pressure to remove the solvent. The resulting crude product was purified by silica gel column chromatography to afford compound 22 as an orange solid (0.52 g, 64\% yield). \textsuperscript{1}H NMR (500 MHz, DMSO-d\textsubscript{6}) $\delta$ 10.37 (s, 1H), 8.74 (dd, J = 8.4, 1.5 Hz, 1H), 8.12 (d, J = 2.3 Hz, 1H), 7.96 -- 7.92 (m, 2H), 7.85 (d, J = 16.1 Hz, 1H), 7.81 (dd, J = 8.4, 1.3 Hz, 1H), 7.76 (dd, J = 8.7, 2.4 Hz, 1H), 7.64 -- 7.61 (m, 2H), 7.35 -- 7.30 (m, 1H), 7.25 (d, J = 8.7 Hz, 1H), 7.04 (s, 1H), 7.01 (d, J = 8.5 Hz, 1H), 6.99 -- 6.95 (m, 1H), 6.71 (d, J = 16.1 Hz, 1H), 5.03 (d, J = 2.5 Hz, 2H), 3.67 (t, J = 2.3 Hz, 1H), 1.50 (s, 9H).

\begin{center}\includegraphics[width=0.61\textwidth]{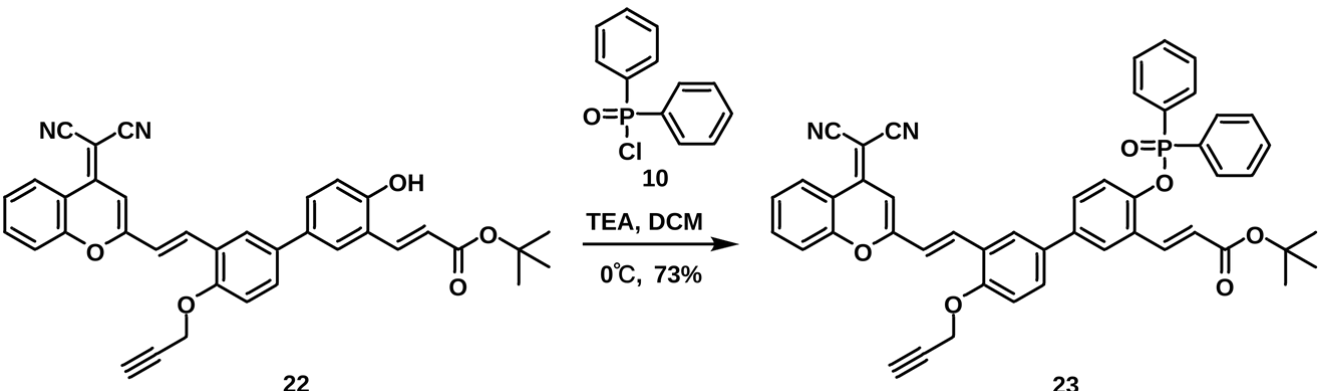}\end{center}\nopagebreak[4]
\vspace{-6pt}\noindent Compound 22 (0.50 g, 0.88 mmol) was dissolved in dichloromethane (DCM, 10 mL). The reaction system was cooled to 0~\textdegree C in an ice bath, and TEA (0.23 mL, 1.32 mmol) was added dropwise slowly. After stirring for 10 min, compound 10 (0.24 g, 1.06 mmol) was added dropwise to the mixture. The reaction progress was monitored by TLC. Upon completion, the reaction was quenched by the slow addition of dilute hydrochloric acid. The organic phase was separated, and the aqueous phase was extracted with DCM three times. The combined organic phases were washed sequentially with deionized water and saturated aqueous sodium chloride solution three times each. The organic phase was concentrated under reduced pressure to remove the solvent, and the crude product was purified by silica gel column chromatography to afford compound 23 as an orange solid (0.49 g, 73\% yield). \textsuperscript{1}H NMR (500 MHz, Chloroform-d) $\delta$ 8.93 (dd, J = 8.4, 1.4 Hz, 1H), 8.05 -- 7.86 (m, 6H), 7.76 (ddd, J = 8.5, 7.1, 1.4 Hz, 1H), 7.70 (dd, J = 4.7, 2.3 Hz, 2H), 7.66 -- 7.56 (m, 4H), 7.55 -- 7.42 (m, 7H), 7.17 (d, J = 8.6 Hz, 1H), 7.06 (d, J = 16.1 Hz, 1H), 6.91 (s, 1H), 6.40 (d, J = 16.1 Hz, 1H), 4.92 (d, J = 2.4 Hz, 2H), 2.64 (t, J = 2.4 Hz, 1H), 1.56 (s, 9H).

\begin{center}\includegraphics[width=0.58\textwidth]{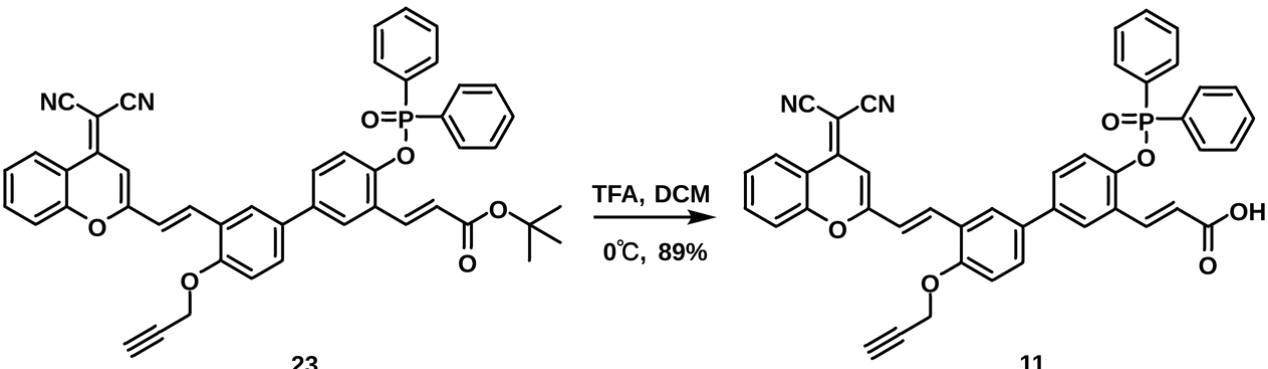}\end{center}\nopagebreak[4]
\vspace{-6pt}\noindent Compound 23 (0.46 g, 0.60 mmol) was dissolved in DCM (8 mL) and cooled to 0~\textdegree C in an ice bath. TFA (2 mL) was added dropwise slowly with stirring. The reaction progress was monitored by TLC. Upon complete consumption of the starting material, the reaction mixture was concentrated under reduced pressure to remove volatile solvents, affording crude compound 11 as an orange solid (0.38 g, 89\% yield).

\clearpage
\begin{center}\includegraphics[width=0.76\textwidth]{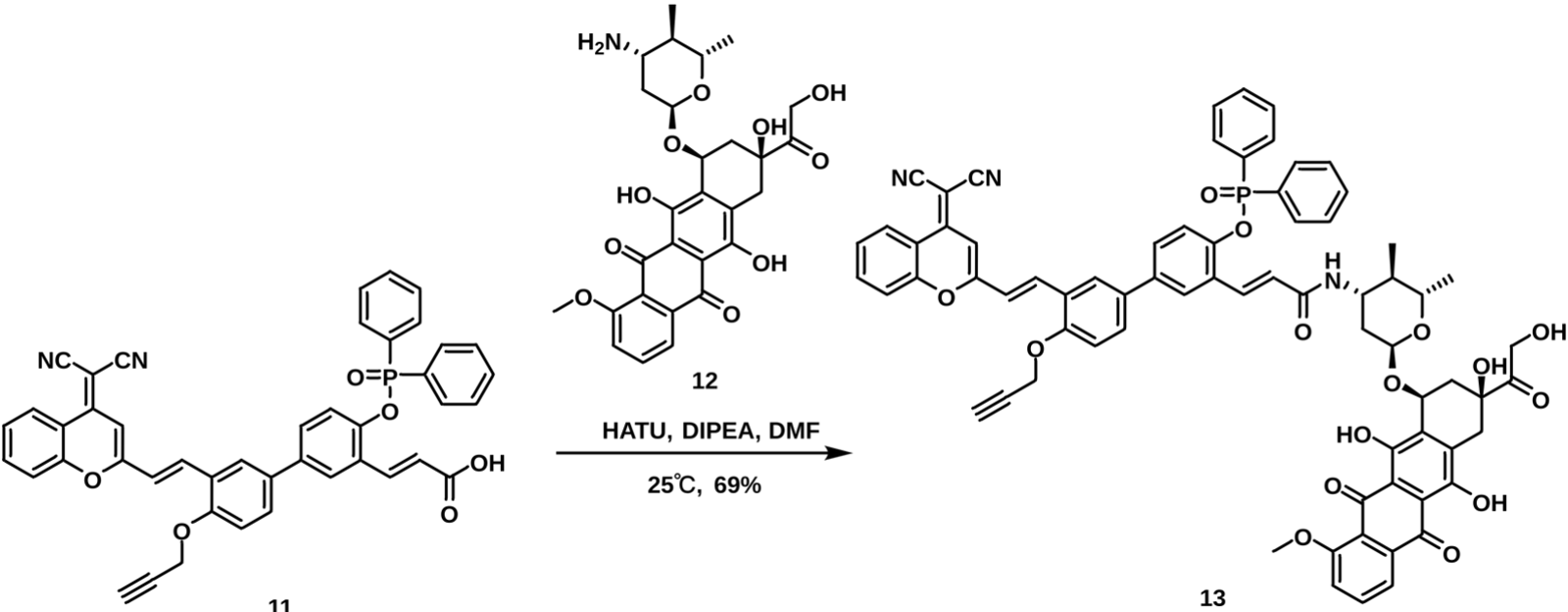}\end{center}\nopagebreak[4]
\vspace{-6pt}\noindent Compound 11 (0.35 g, 0.49 mmol) and doxorubicin (compound 12, 0.32 g, 0.59 mmol) were dissolved in N,N-dimethylformamide (DMF, 10 mL), and the reaction system was cooled in an ice bath. N,N-Diisopropylethylamine (DIPEA, 0.17 mL, 0.98 mmol) was added dropwise slowly with stirring. After stirring for 10 min, 2-(7-aza-1H-benzotriazol-1-yl)-1,1,3,3-tetramethyluronium hexafluorophosphate (HATU, 0.56 g, 1.47 mmol) was added slowly in portions. The reaction was allowed to proceed at 25~\textdegree C for another 20 min. After completion, the reaction was quenched by slow dropwise addition of deionized water. After full precipitation of the solid, the mixture was filtered under reduced pressure, and the filter cake was collected as the crude product. Purification by silica gel column chromatography afforded compound 13 as a dark red solid (0.42 g, 69\% yield). \textsuperscript{1}H NMR (500 MHz, DMSO-d\textsubscript{6}) $\delta$ 14.00 (s, 1H), 13.23 (s, 1H), 8.66 (dd, J = 8.3, 1.4 Hz, 1H), 8.02 (d, J = 2.2 Hz, 1H), 7.91 -- 7.84 (m, 5H), 7.84 -- 7.76 (m, 6H), 7.73 -- 7.69 (m, 1H), 7.66 -- 7.58 (m, 3H), 7.57 -- 7.49 (m, 8H), 7.39 -- 7.33 (m, 1H), 7.25 (d, J = 8.8 Hz, 1H), 6.99 -- 6.89 (m, 2H), 5.46 (s, 1H), 5.26 (s, 1H), 5.03 (d, J = 2.2 Hz, 2H), 4.97 -- 4.85 (m, 3H), 4.62 (d, J = 6.2 Hz, 2H), 4.24 (d, J = 6.5 Hz, 1H), 4.18 (s, 1H), 3.91 (s, 3H), 3.68 (t, J = 2.3 Hz, 1H), 3.51 -- 3.48 (m, 1H), 2.99 (d, J = 17.9 Hz, 1H), 2.90 (d, J = 12.3 Hz, 1H), 2.24 (d, J = 13.9 Hz, 1H), 2.10 (d, J = 10.0 Hz, 1H), 1.91 (d, J = 22.4 Hz, 1H), 1.58 (d, J = 8.2 Hz, 1H), 1.17 (d, J = 6.4 Hz, 3H).

\begin{center}\includegraphics[width=0.48\textwidth]{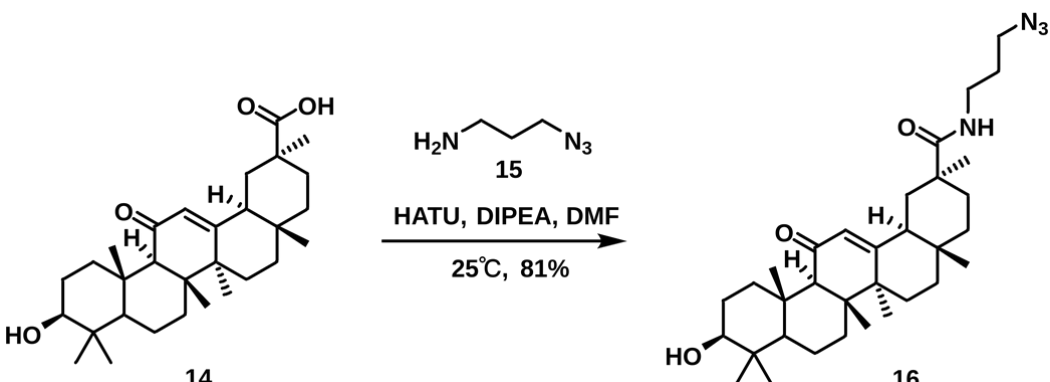}\end{center}\nopagebreak[4]
\vspace{-6pt}\noindent Compound 14 (0.37 g, 0.79 mmol) and compound 15 (0.10 g, 1.02 mmol) were dissolved in DMF (10 mL), and the reaction system was cooled in an ice bath. DIPEA (0.41 mL, 2.36 mmol) was added dropwise slowly with stirring. After stirring for 10 min, HATU (0.60 g, 1.57 mmol) was added slowly in portions. The reaction was allowed to proceed at 25~\textdegree C for another 20 min. After completion, the reaction was quenched by slow dropwise addition of deionized water. After full precipitation of the solid, the mixture was filtered under reduced pressure, and the filter cake was collected as the crude product. Purification by silica gel column chromatography afforded compound 16 as a white solid (0.35 g, 81\% yield). \textsuperscript{1}H NMR (500 MHz, DMSO-d\textsubscript{6}) $\delta$ 7.66 (t, J = 5.7 Hz, 1H), 5.49 (s, 1H), 4.31 (s, 1H), 3.94 (s, 1H), 3.62 (t, J = 6.6 Hz, 2H), 3.23 (dq, J = 12.7, 6.4 Hz, 1H), 3.19 -- 3.13 (m, 1H), 3.02 (dd, J = 11.6, 4.5 Hz, 1H), 2.59 (dt, J = 13.3, 3.4 Hz, 1H), 2.31 (s, 1H), 2.08 (td, J = 13.6, 4.6 Hz, 1H), 2.04 -- 2.01 (m, 1H), 1.92 -- 1.85 (m, 3H), 1.81 (dt, J = 13.7, 3.6 Hz, 1H), 1.74 (td, J = 13.7, 4.8 Hz, 1H), 1.66 -- 1.59 (m, 2H), 1.52 (ddd, J = 11.7, 8.7, 5.4 Hz, 2H), 1.41 (dt, J = 12.6, 3.6 Hz, 1H), 1.38 -- 1.36 (m, 1H), 1.35 (s, 3H), 1.32 (dd, J = 11.4, 3.5 Hz, 1H), 1.29 -- 1.25 (m, 2H), 1.20 -- 1.10 (m, 2H), 1.03 (s, 6H), 1.02 (s, 3H), 0.95 (dt, J = 13.6, 4.1 Hz, 2H), 0.91 (s, 3H), 0.73 (s, 3H), 0.69 (s, 3H).

\clearpage
\begin{center}\includegraphics[width=0.92\textwidth]{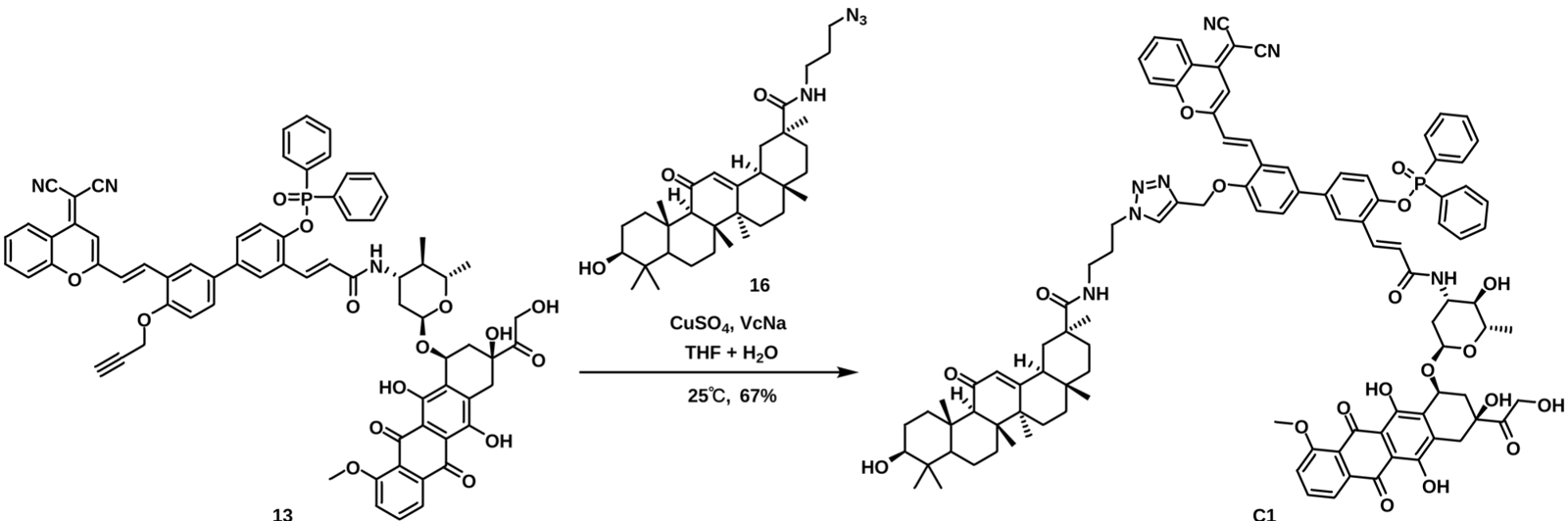}\end{center}\nopagebreak[4]
\vspace{-6pt}\noindent Compound 13 (0.40 g, 0.32 mmol) and compound 16 (0.21 g, 0.39 mmol) were dissolved in tetrahydrofuran (THF, 8 mL) and stirred until homogeneous for further use. Separately, copper sulfate (CuSO$_4$, 0.05 g, 0.25 mmol) and sodium ascorbate (VcNa, 0.03 g, 0.19 mmol) were fully dissolved in deionized water (2 mL) to prepare an aqueous catalyst solution. The aqueous catalyst solution was added to the above reaction mixture in portions. After purging the system with nitrogen to displace air, the reaction was stirred at 25~\textdegree C under a nitrogen atmosphere for 3 h. Upon complete consumption of starting materials as monitored by TLC, the mixture was concentrated under reduced pressure to remove the organic solvent, and the precipitated crude solid was collected. The crude product was purified by silica gel column chromatography to afford compound C1 as a dark red solid (0.39 g, 67\% yield). \textsuperscript{1}H NMR (500 MHz, DMSO-d\textsubscript{6}) $\delta$ 14.02 (s, 1H), 13.24 (s, 1H), 8.66 (d, J = 8.3 Hz, 1H), 8.35 (s, 1H), 8.04 (s, 1H), 7.96 -- 7.83 (m, 9H), 7.82 -- 7.78 (m, 2H), 7.71 -- 7.55 (m, 8H), 7.55 -- 7.50 (m, 5H), 7.37 (dd, J = 16.9, 8.7 Hz, 2H), 6.94 (d, J = 15.7 Hz, 1H), 6.88 (s, 1H), 5.46 (s, 1H), 5.38 (d, J = 8.6 Hz, 3H), 5.27 (s, 1H), 4.96 -- 4.86 (m, 3H), 4.62 (s, 2H), 4.40 (t, J = 7.0 Hz, 2H), 4.29 (s, 1H), 4.26 -- 4.21 (m, 1H), 4.18 (dt, J = 10.9, 5.7 Hz, 1H), 3.93 (d, J = 10.0 Hz, 4H), 3.50 (s, 1H), 3.13 (q, J = 6.6 Hz, 2H), 2.97 (t, J = 20.0 Hz, 3H), 2.23 (s, 2H), 2.11 (d, J = 9.0 Hz, 1H), 2.03 -- 1.96 (m, 4H), 1.94 -- 1.83 (m, 2H), 1.76 (d, J = 13.0 Hz, 1H), 1.67 -- 1.36 (m, 8H), 1.27 (s, 5H), 1.23 (d, J = 6.0 Hz, 3H), 1.17 (dd, J = 5.1, 3.2 Hz, 3H), 0.99 (s, 3H), 0.96 (s, 3H), 0.92 (s, 3H), 0.87 (s, 5H), 0.65 (d, J = 9.5 Hz, 6H). \textsuperscript{13}C NMR (126 MHz, DMSO-d\textsubscript{6}) $\delta$ 214.30, 212.55, 199.44, 186.84, 186.55, 175.91, 170.02, 164.38, 162.80, 161.15, 158.65, 156.66, 155.00, 153.09, 152.26, 148.74, 145.78, 142.92, 142.31, 141.87, 139.86, 139.42, 139.08, 137.77, 136.73, 135.88, 134.94, 134.48, 133.43, 132.98, 132.52, 131.73, 131.35, 131.29, 130.38, 129.55, 129.46, 127.88, 127.33, 127.17, 126.63, 126.52, 126.41, 125.52, 125.42, 124.95, 124.42, 124.16, 123.29, 122.04, 121.20, 120.50, 120.28, 119.31, 118.68, 117.46, 116.24, 115.14, 111.17, 110.97, 107.13, 100.82, 77.06, 76.81, 75.45, 74.01, 68.65, 66.60, 64.23, 63.88, 61.58, 60.71, 56.90, 54.54, 48.11, 47.91, 45.14, 43.34, 43.23, 39.18, 38.97, 37.79, 37.03, 36.39, 36.26, 32.51, 31.75, 31.24, 30.72, 30.29, 29.49, 29.10, 28.79, 28.54, 27.37, 26.32, 25.41, 23.37, 18.67, 17.51, 16.53, 16.39. HRMS: m/z [M+2H]\textsuperscript{2+} calcd 897.8809, found 897.8825.

\clearpage
\subsection*{Compound C2}

\begin{center}\includegraphics[width=0.49\textwidth]{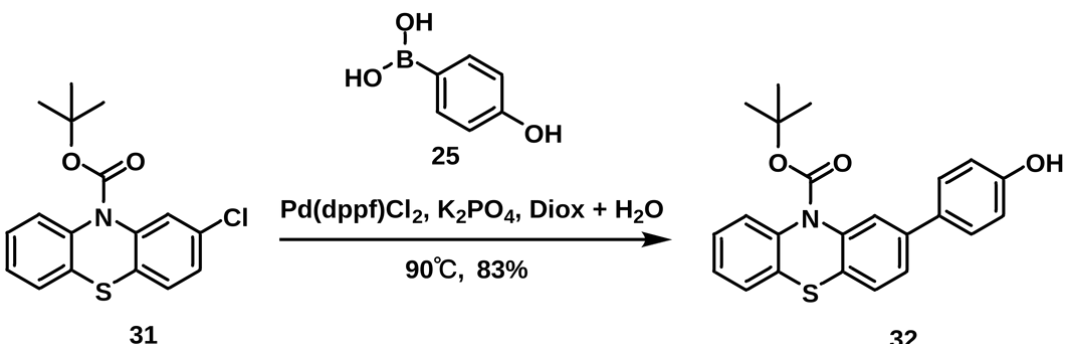}\end{center}\nopagebreak[4]
\vspace{-6pt}\noindent Compound 31 (0.47 g, 1.41 mmol), compound 25 (0.29 g, 2.11 mmol), Pd(dppf)Cl\textsubscript{2} (0.05 g, 0.07 mmol), and K\textsubscript{3}PO\textsubscript{4} (0.60 g, 2.82 mmol) were sequentially charged into a side-arm ground-glass Schlenk flask. Diox (8 mL) and deionized water (2 mL) were added as the mixed reaction solvent. The system was evacuated and backfilled with nitrogen three times via a Schlenk manifold, and the reaction mixture was stirred at 90~\textdegree C for 16 h. Upon completion, the mixture was filtered through a Celite pad to remove insoluble residues. The filtrate was concentrated under reduced pressure, and the crude product was purified by silica gel column chromatography to afford compound 32 as a white solid (0.46 g, 83\% yield). \textsuperscript{1}H NMR (500 MHz, Chloroform-d) $\delta$ 7.74 (d, J = 1.9 Hz, 1H), 7.55 (dd, J = 8.1, 1.3 Hz, 1H), 7.44 -- 7.37 (m, 4H), 7.35 (dd, J = 8.1, 1.9 Hz, 1H), 7.32 -- 7.28 (m, 1H), 7.19 (td, J = 7.6, 1.3 Hz, 1H), 6.91 -- 6.85 (m, 2H), 1.53 (s, 9H).

\begin{center}\includegraphics[width=0.48\textwidth]{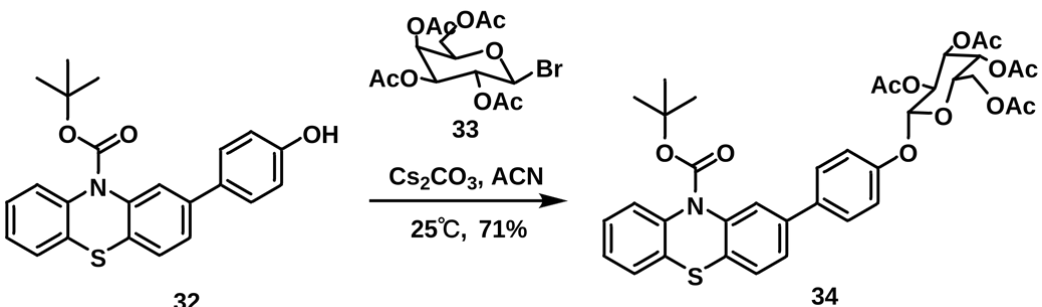}\end{center}\nopagebreak[4]
\vspace{-6pt}\noindent Compound 32 (0.41 g, 1.05 mmol) was dissolved in acetonitrile (ACN, 10 mL). After stirring until complete dissolution, Cs\textsubscript{2}CO\textsubscript{3} (0.68 g, 2.09 mmol) was added, and the mixture was stirred at 25~\textdegree C (room temperature) for 10 min. Compound 33 (0.43 g, 1.05 mmol) was then added to the reaction system, and the mixture was stirred at a constant temperature of 25~\textdegree C for 3 h. Upon complete consumption of the starting material as monitored by thin-layer chromatography (TLC), the reaction mixture was filtered through a Celite pad to remove insoluble inorganic salts. The filtrate was concentrated under reduced pressure to remove the solvent, and the crude product was purified by silica gel column chromatography to afford compound 34 as a white viscous syrup (0.54 g, 71\% yield). \textsuperscript{1}H NMR (500 MHz, Chloroform-d) $\delta$ 7.73 (d, J = 1.9 Hz, 1H), 7.54 (dd, J = 8.2, 1.3 Hz, 1H), 7.53 -- 7.51 (m, 2H), 7.40 (dd, J = 8.1, 1.6 Hz, 1H), 7.37 (dt, J = 7.8, 1.6 Hz, 1H), 7.34 (ddd, J = 8.1, 2.0, 0.8 Hz, 1H), 7.29 (ddd, J = 8.1, 5.9, 1.4 Hz, 1H), 7.17 (tt, J = 7.5, 1.5 Hz, 1H), 7.10 -- 7.07 (m, 2H), 5.54 (dd, J = 10.5, 7.9 Hz, 1H), 5.49 (dd, J = 3.5, 1.1 Hz, 1H), 5.16 (ddd, J = 10.3, 3.4, 1.0 Hz, 1H), 5.12 (dd, J = 8.0, 1.1 Hz, 1H), 4.27 (dd, J = 11.3, 6.9 Hz, 1H), 4.20 (dd, J = 11.4, 6.3 Hz, 1H), 4.12 (td, J = 6.7, 1.2 Hz, 1H), 2.21 (d, J = 1.2 Hz, 3H), 2.11 (s, 3H), 2.08 (d, J = 1.0 Hz, 3H), 2.04 (s, 3H), 1.52 (s, 9H).

\clearpage
\begin{center}\includegraphics[width=0.53\textwidth]{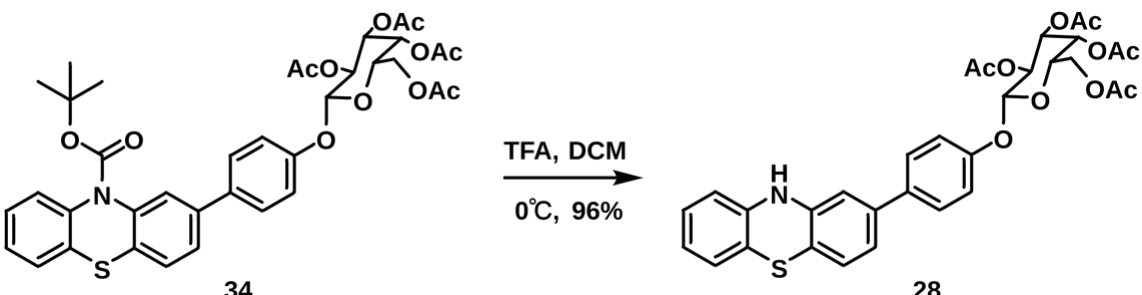}\end{center}\nopagebreak[4]
\vspace{-6pt}\noindent Compound 34 (0.46 g, 0.64 mmol) was dissolved in DCM (8 mL) and cooled to 0~\textdegree C in an ice bath. TFA (2 mL) was added dropwise with stirring. The reaction progress was monitored by TLC. Upon complete consumption of the starting material, the mixture was concentrated under reduced pressure to remove volatile components, yielding crude compound 28 as a white syrup (0.38 g, 96\% yield). \textsuperscript{1}H NMR (500 MHz, DMSO-d\textsubscript{6}) $\delta$ 8.67 (s, 1H), 7.57 -- 7.50 (m, 2H), 7.10 -- 7.05 (m, 2H), 7.03 -- 6.96 (m, 3H), 6.95 -- 6.90 (m, 2H), 6.77 (td, J = 7.5, 1.3 Hz, 1H), 6.70 (dd, J = 8.0, 1.3 Hz, 1H), 5.52 (d, J = 7.8 Hz, 1H), 5.36 (dd, J = 3.4, 1.0 Hz, 1H), 5.31 (dd, J = 10.4, 3.5 Hz, 1H), 5.24 (dd, J = 10.3, 7.8 Hz, 1H), 4.49 -- 4.43 (m, 1H), 4.15 -- 4.09 (m, 2H), 2.17 (s, 3H), 2.05 (d, J = 14.2 Hz, 6H), 1.96 (s, 3H).

\begin{center}\includegraphics[width=0.92\textwidth]{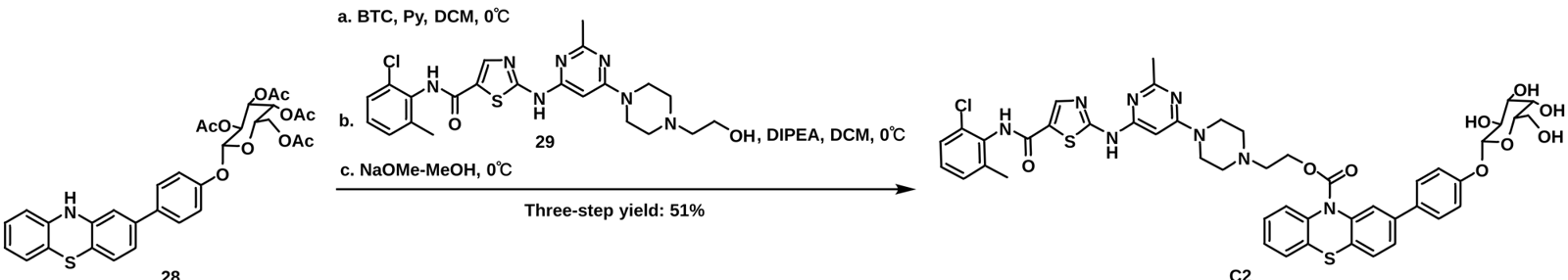}\end{center}\nopagebreak[4]
\vspace{-6pt}\noindent Compound 28 (0.31 g, 0.50 mmol) was dissolved in DCM (6 mL) and cooled to 0~\textdegree C in an ice bath. Pyridine (Py, 0.12 mL, 0.15 mmol) was added dropwise, and the mixture was stirred at this temperature for 10 min. Then, a solution of BTC (0.15 g, 0.50 mmol) in DCM (2 mL) was added slowly dropwise, and the reaction was maintained at 0~\textdegree C for 2 h. Subsequently, a solution of dasatinib (compound 29, 0.24 g, 0.50 mmol) in DCM (2 mL) was added dropwise to the mixture. After thorough mixing, N,N-diisopropylethylamine (DIPEA, 0.17 mL, 1.00 mmol) was added dropwise, and the reaction was stirred at 0~\textdegree C for another 2 h. Upon completion as monitored by TLC, 3 mL of 4 M sodium methoxide in methanol (NaOMe-MeOH) was added, and the mixture was stirred at 0~\textdegree C for 1 h to remove the acetyl protecting groups on the glycosyl moiety. After the reaction, an appropriate amount of cation exchange resin was added, and the mixture was stirred for 10 min until the solution became clear to neutralize excess base. The mixture was filtered through a Celite pad. The filtrate was concentrated under reduced pressure, and the crude product was purified by silica gel column chromatography to afford the target product C2 as a white solid (0.25 g, 51\% yield). \textsuperscript{1}H NMR (500 MHz, DMSO-d\textsubscript{6}) $\delta$ 10.20 (s, 1H), 8.73 (s, 1H), 7.58 (s, 1H), 7.52 -- 7.38 (m, 6H), 7.31 (dt, J = 15.5, 7.4 Hz, 3H), 7.05 (d, J = 8.3 Hz, 2H), 6.16 (s, 1H), 5.20 (dd, J = 8.2, 5.2 Hz, 1H), 4.89 -- 4.82 (m, 2H), 4.66 (dt, J = 20.7, 5.0 Hz, 1H), 4.53 (d, J = 4.7 Hz, 1H), 4.48 (t, J = 5.3 Hz, 1H), 3.74 (dt, J = 5.2, 1.9 Hz, 1H), 3.61 -- 3.55 (m, 8H), 3.51 (t, J = 6.0 Hz, 1H), 3.43 (dt, J = 6.0, 2.2 Hz, 1H), 2.46 (t, J = 6.3 Hz, 2H), 2.36 -- 2.21 (m, 6H). \textsuperscript{13}C NMR (126 MHz, MeOD) $\delta$ 176.35, 166.97, 163.45, 162.13, 157.69, 155.65, 154.86, 146.62, 145.27, 144.49, 143.34, 140.33, 138.80, 137.40, 136.83, 135.83, 135.08, 133.24, 132.64, 130.48, 128.88, 128.41, 127.63, 127.01, 125.64, 122.59, 121.97, 119.19, 118.19, 116.76, 101.33, 92.70, 75.50, 73.46, 70.86, 68.74, 60.90, 58.98, 56.47, 51.98, 42.11, 33.52, 24.31. HRMS: m/z [M+H]\textsuperscript{+} calcd 967.2669, found 967.2656.

\clearpage

\begin{figure}[p]
\centering
\includegraphics[width=0.9\textwidth,height=96mm,keepaspectratio]{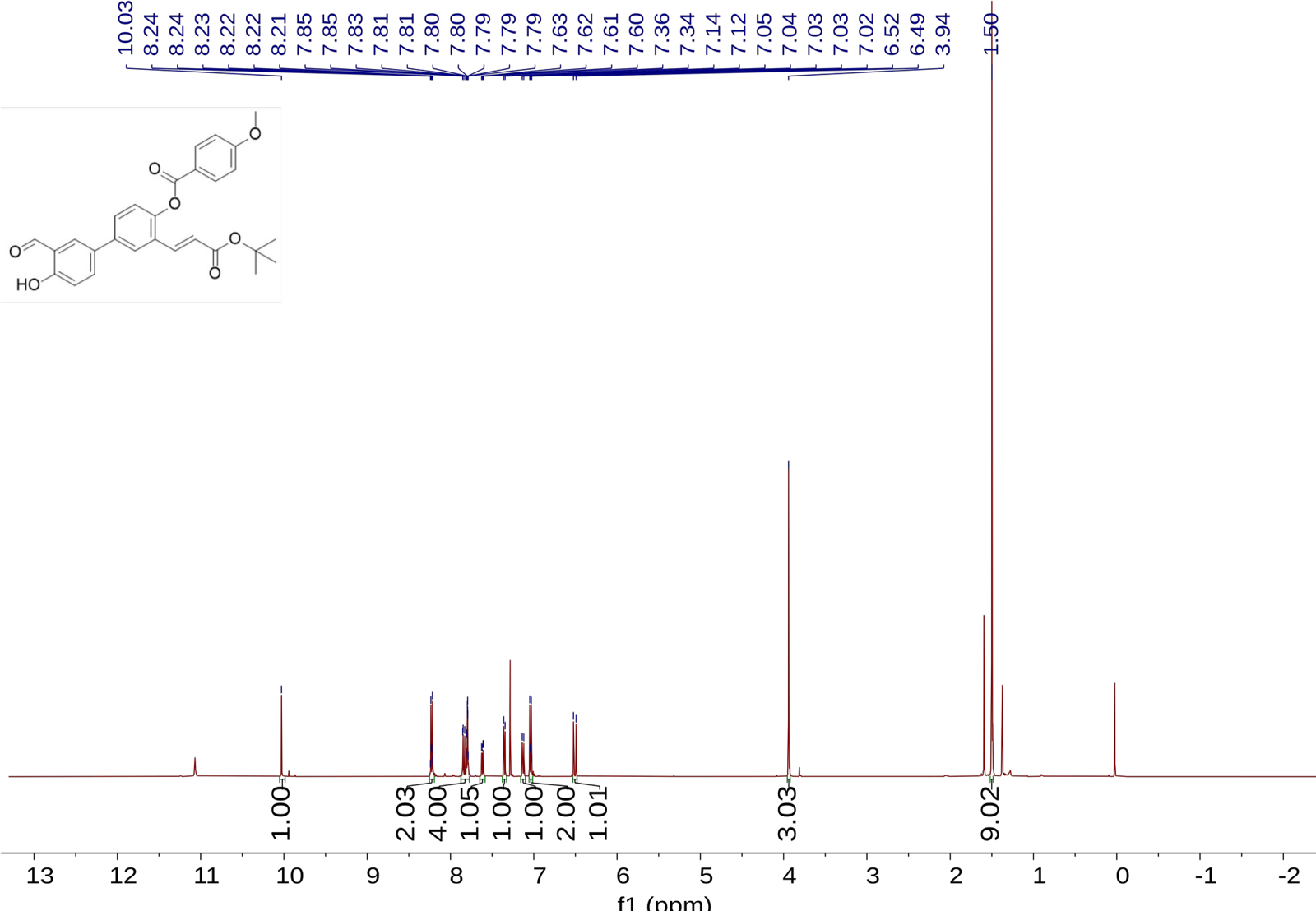}
\par\smallskip
{\small \textbf{a}, compound 19, \textsuperscript{1}H NMR (500 MHz, CDCl\textsubscript{3})\par}
\vspace{5mm}
\includegraphics[width=0.9\textwidth,height=96mm,keepaspectratio]{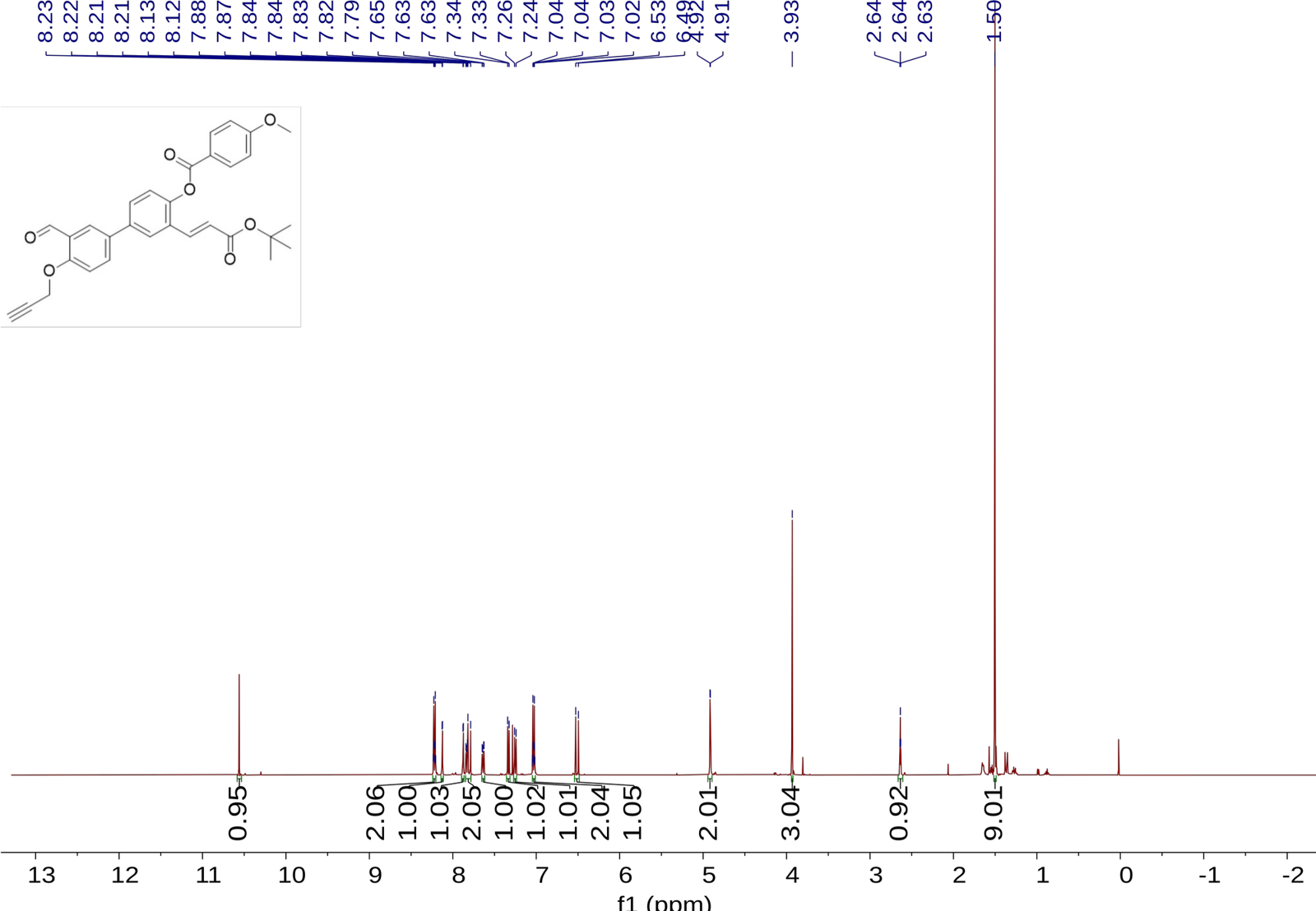}
\par\smallskip
{\small \textbf{b}, compound 21, \textsuperscript{1}H NMR (500 MHz, CDCl\textsubscript{3})\par}
\vspace{5mm}
\caption{\textbf{NMR and HRMS spectra of compound C1 and its intermediates.} Compound numbers follow Figure~\ref{fig:4}; spectra are reproduced as recorded.}
\label{fig:A-specC1}
\end{figure}
\clearpage

\begin{figure}[p]
\ContinuedFloat
\centering
\includegraphics[width=0.9\textwidth,height=96mm,keepaspectratio]{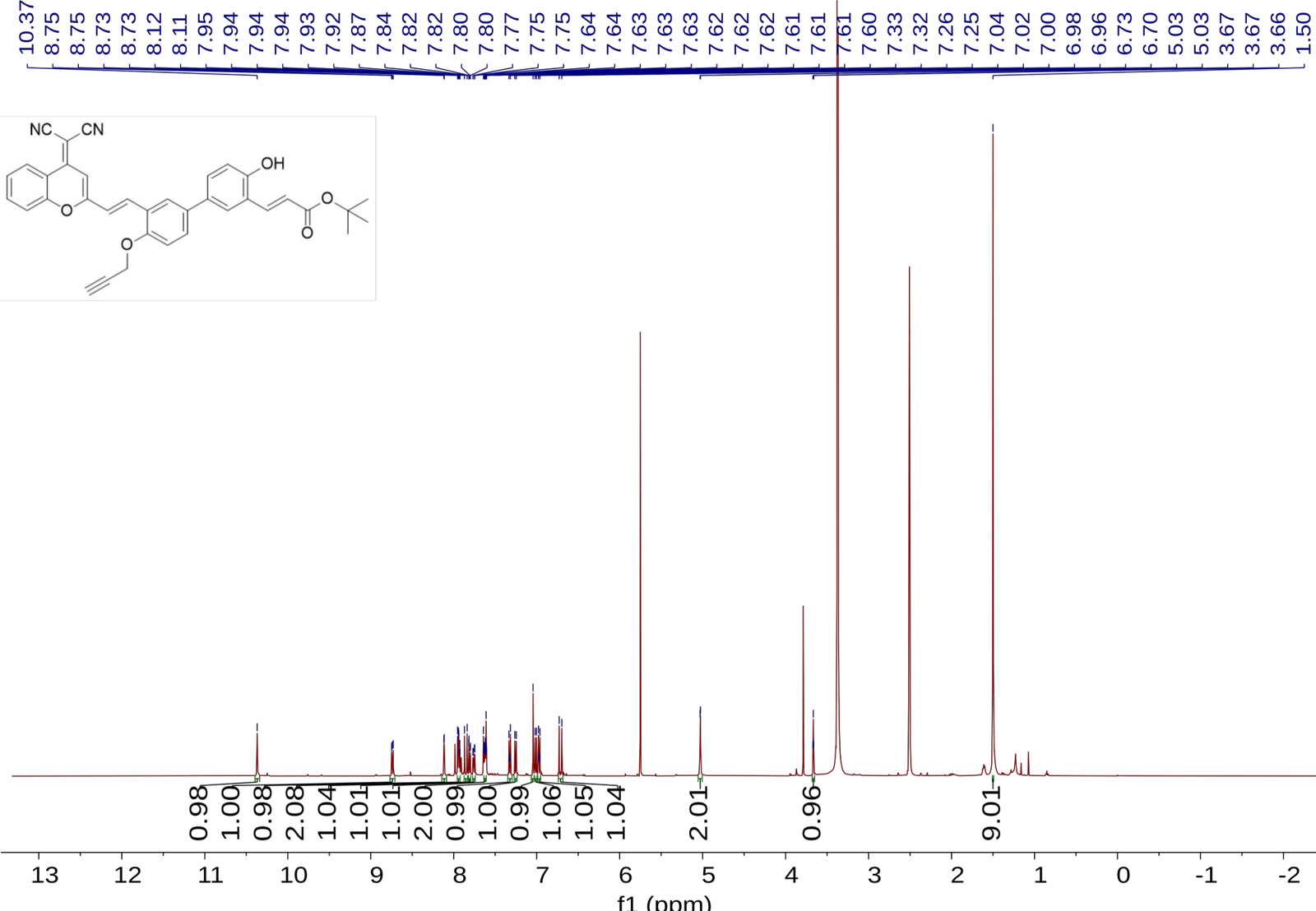}
\par\smallskip
{\small \textbf{c}, compound 22, \textsuperscript{1}H NMR (500 MHz, DMSO-\textit{d}\textsubscript{6})\par}
\vspace{5mm}
\includegraphics[width=0.9\textwidth,height=96mm,keepaspectratio]{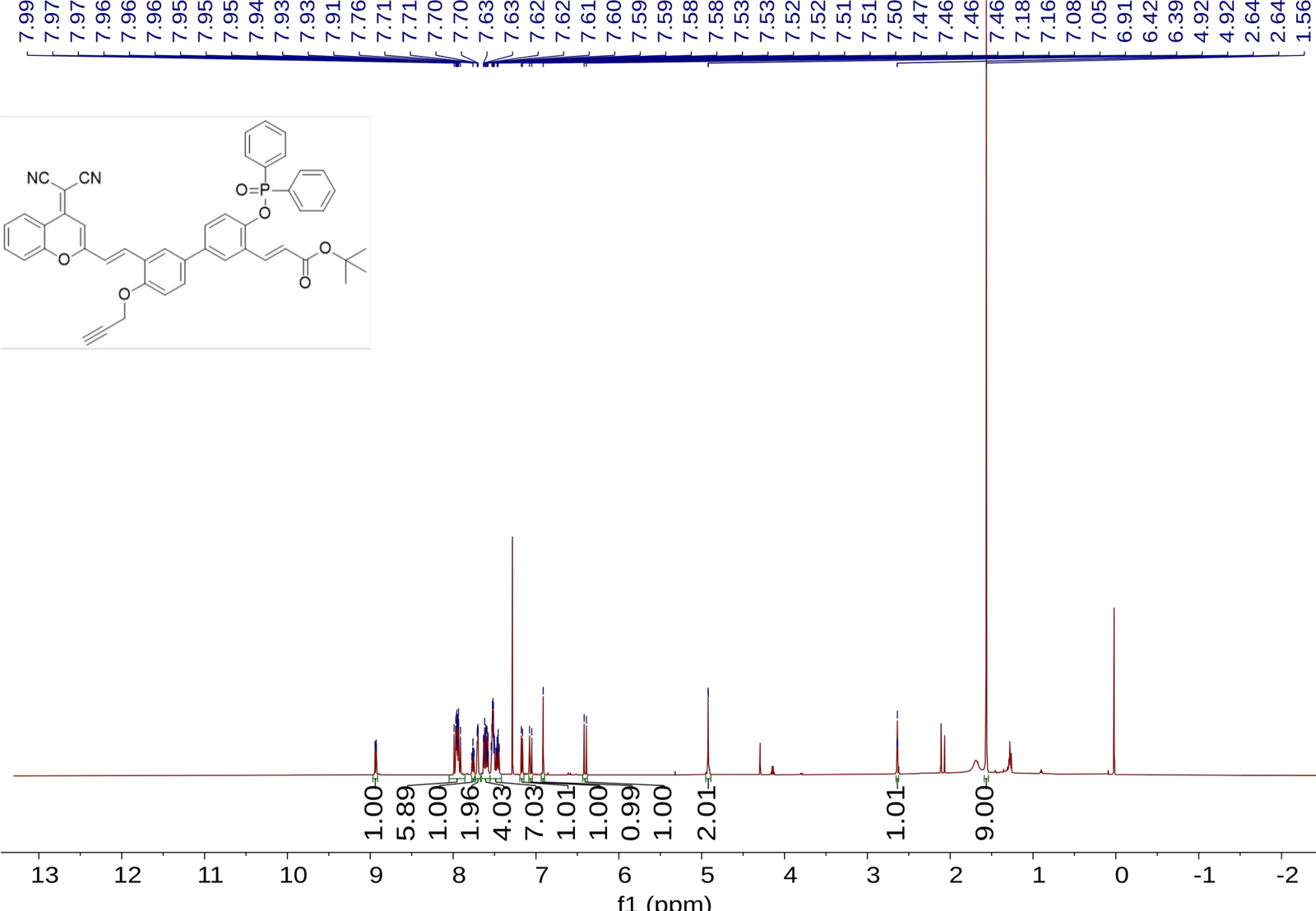}
\par\smallskip
{\small \textbf{d}, compound 23, \textsuperscript{1}H NMR (500 MHz, CDCl\textsubscript{3})\par}
\vspace{5mm}
\caption[]{\textit{Continued.} \textbf{NMR and HRMS spectra of compound C1 and its intermediates.}}
\end{figure}
\clearpage

\begin{figure}[p]
\ContinuedFloat
\centering
\includegraphics[width=0.9\textwidth,height=96mm,keepaspectratio]{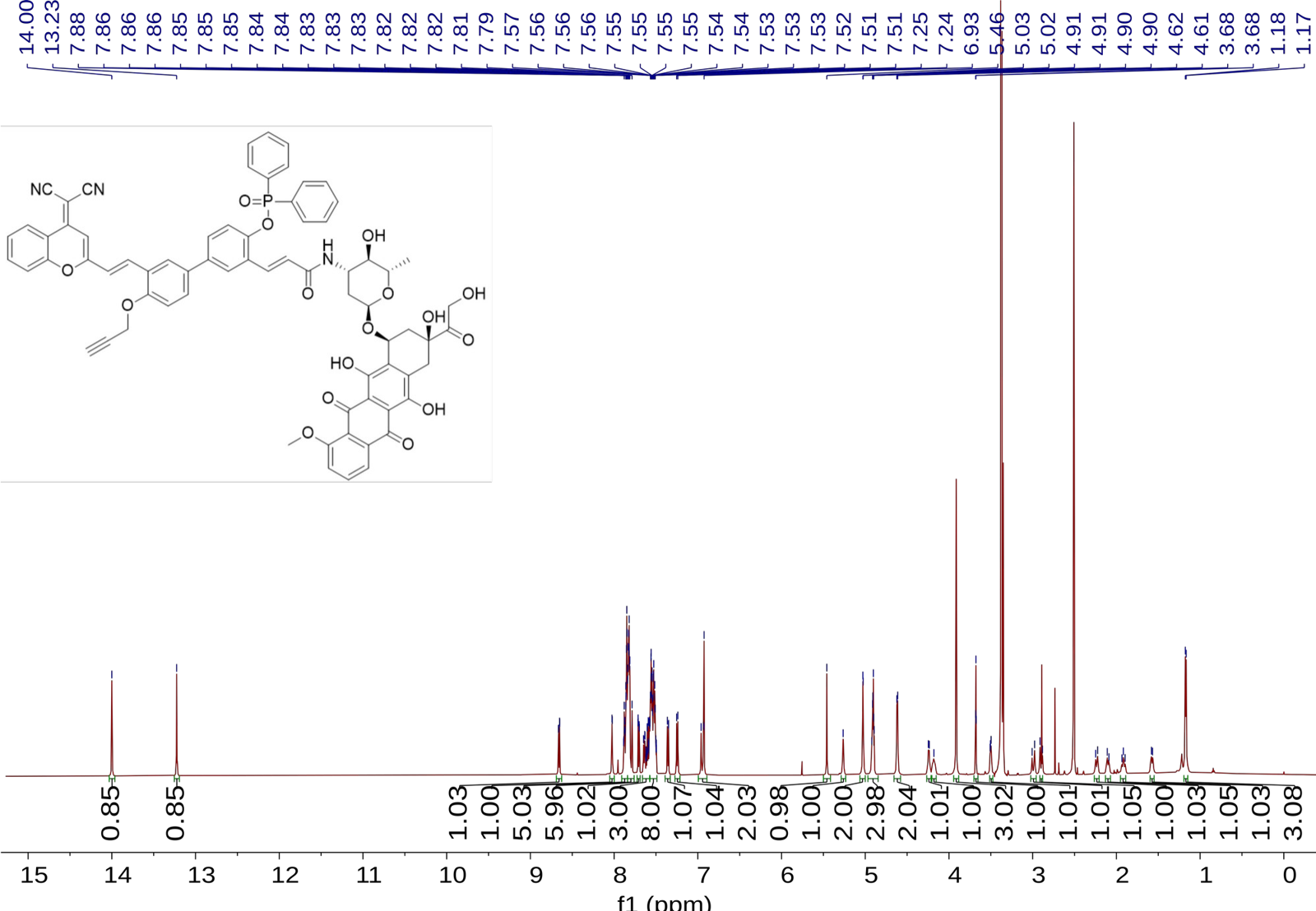}
\par\smallskip
{\small \textbf{e}, compound 13, \textsuperscript{1}H NMR (500 MHz, DMSO-\textit{d}\textsubscript{6})\par}
\vspace{5mm}
\includegraphics[width=0.9\textwidth,height=96mm,keepaspectratio]{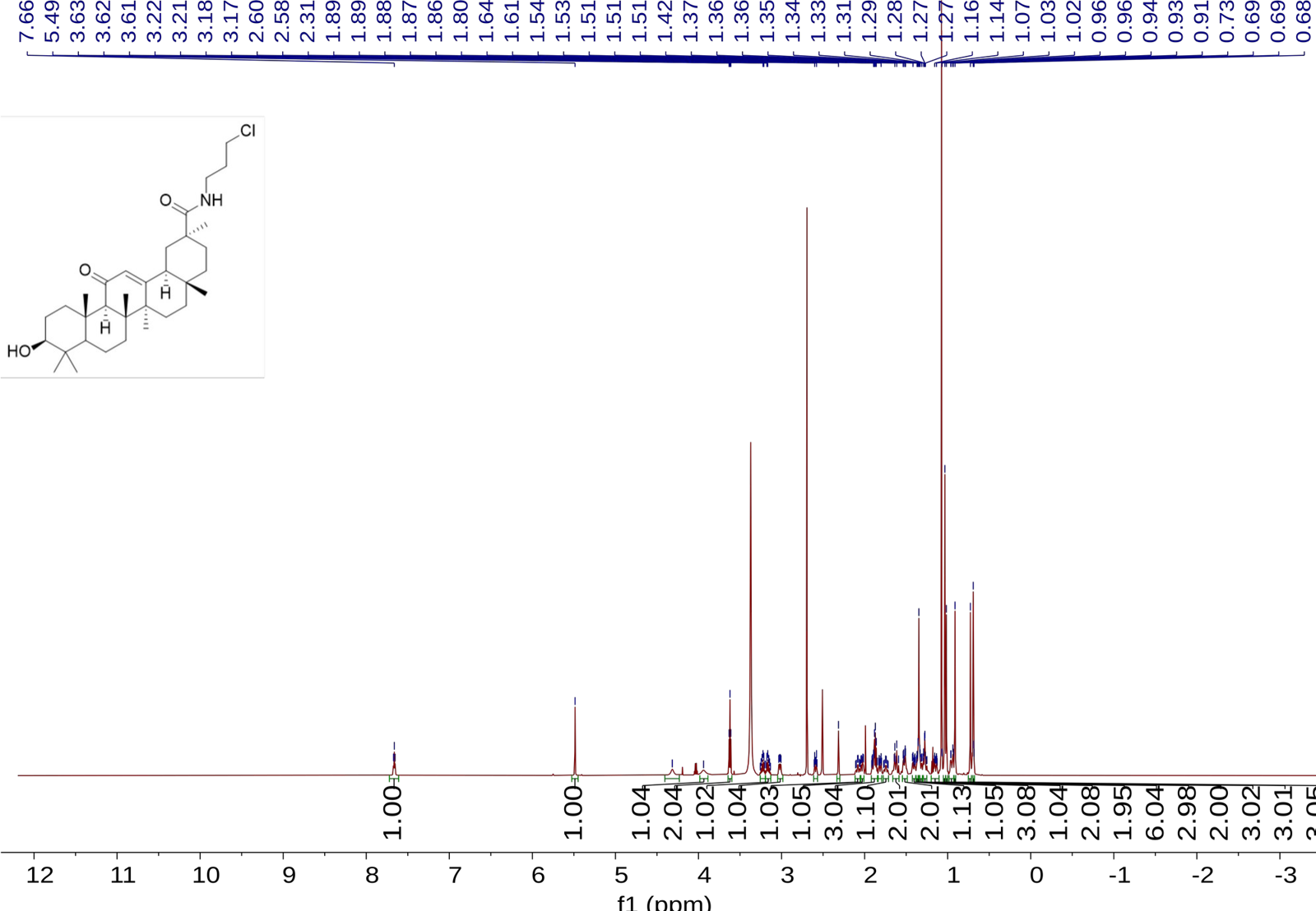}
\par\smallskip
{\small \textbf{f}, compound 16, \textsuperscript{1}H NMR (500 MHz, DMSO-\textit{d}\textsubscript{6})\par}
\vspace{5mm}
\caption[]{\textit{Continued.} \textbf{NMR and HRMS spectra of compound C1 and its intermediates.}}
\end{figure}
\clearpage

\begin{figure}[p]
\ContinuedFloat
\centering
\includegraphics[width=0.9\textwidth,height=96mm,keepaspectratio]{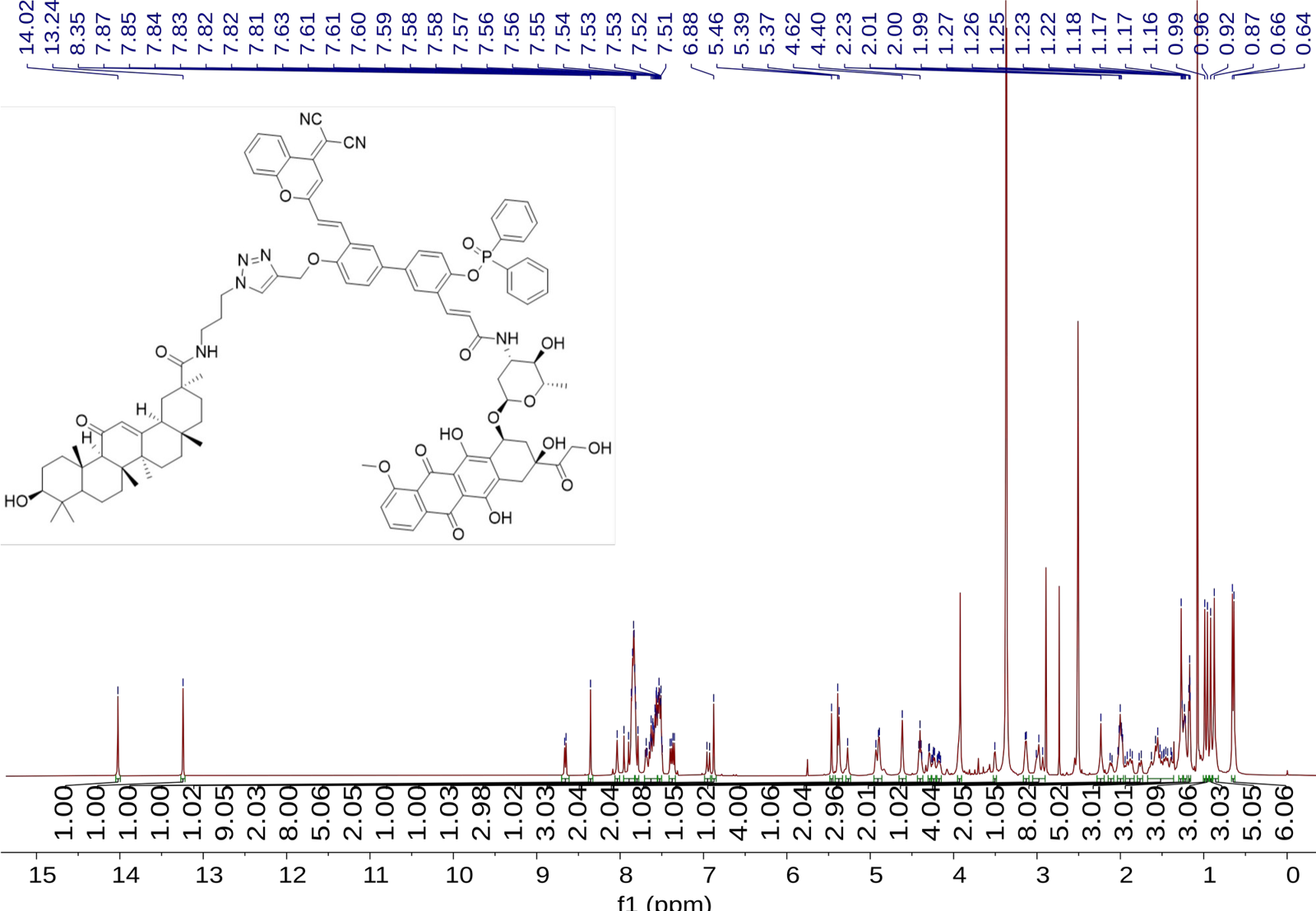}
\par\smallskip
{\small \textbf{g}, C1, \textsuperscript{1}H NMR (500 MHz, DMSO-\textit{d}\textsubscript{6})\par}
\vspace{5mm}
\includegraphics[width=0.9\textwidth,height=96mm,keepaspectratio]{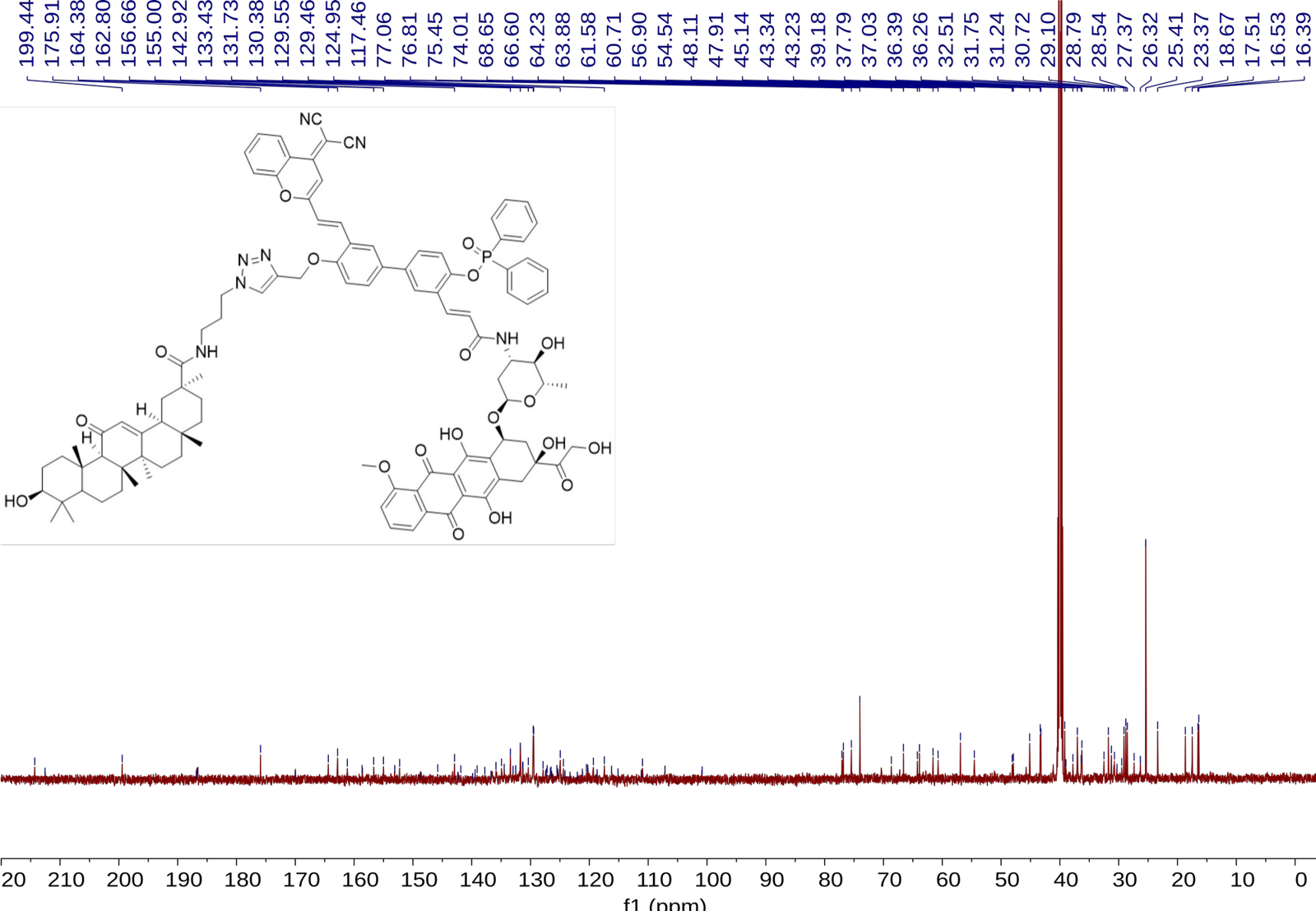}
\par\smallskip
{\small \textbf{h}, C1, \textsuperscript{13}C NMR (126 MHz, DMSO-\textit{d}\textsubscript{6})\par}
\vspace{5mm}
\caption[]{\textit{Continued.} \textbf{NMR and HRMS spectra of compound C1 and its intermediates.}}
\end{figure}
\clearpage

\begin{figure}[p]
\ContinuedFloat
\centering
\includegraphics[width=0.9\textwidth,height=96mm,keepaspectratio]{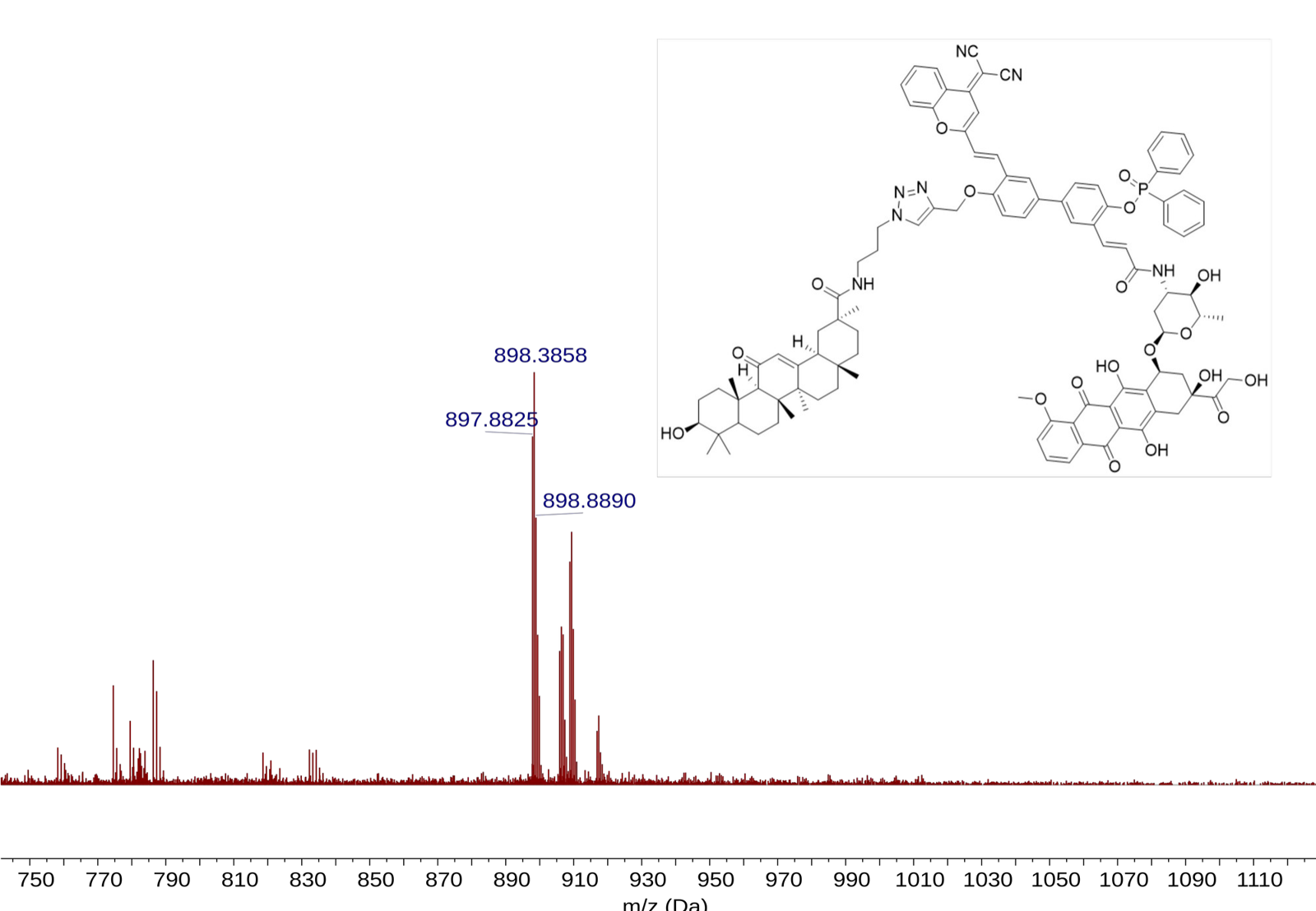}
\par\smallskip
{\small \textbf{i}, C1, HRMS\par}
\vspace{5mm}
\caption[]{\textit{Continued.} \textbf{NMR and HRMS spectra of compound C1 and its intermediates.}}
\end{figure}
\clearpage

\begin{figure}[p]
\centering
\includegraphics[width=0.9\textwidth,height=96mm,keepaspectratio]{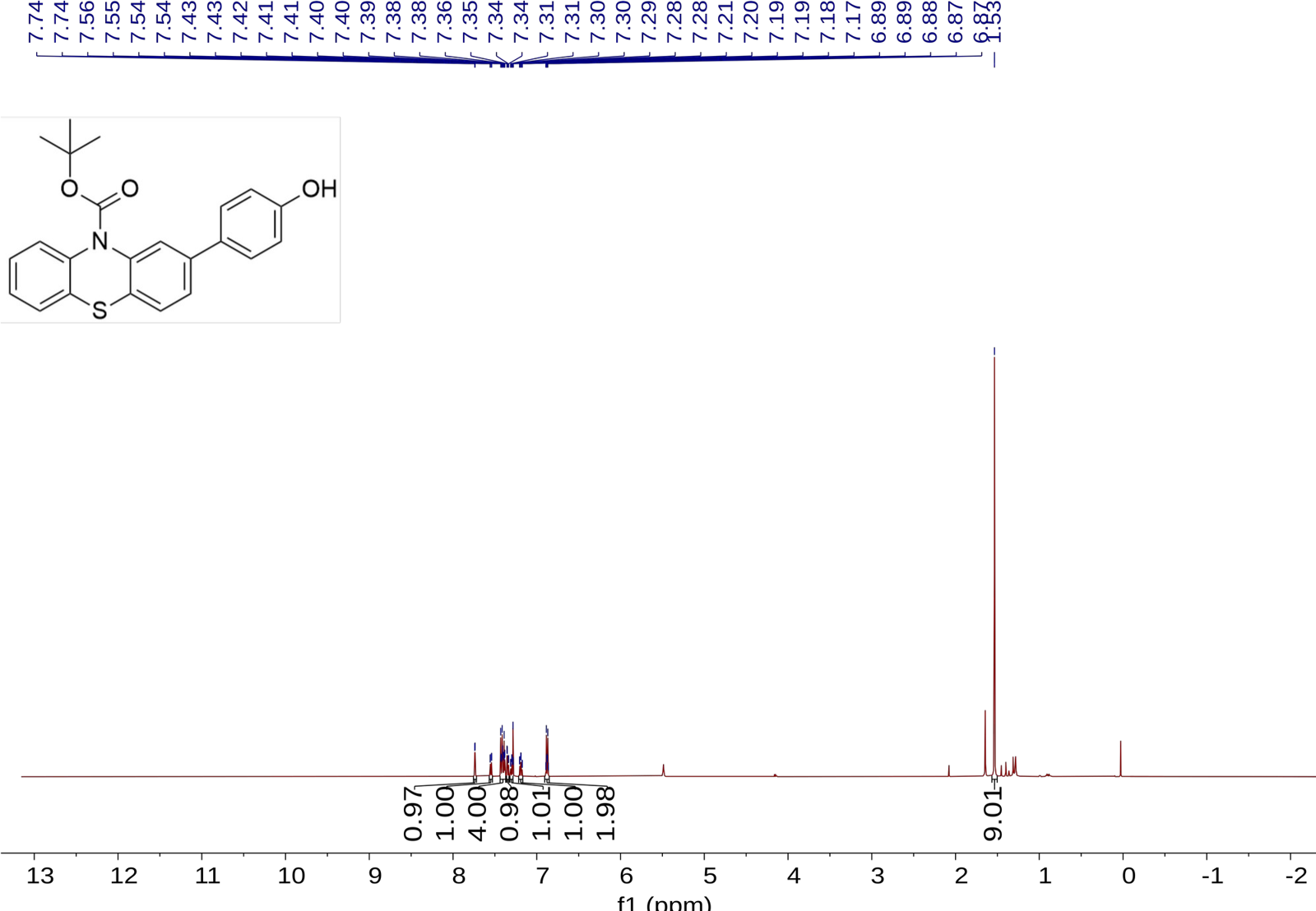}
\par\smallskip
{\small \textbf{a}, compound 32, \textsuperscript{1}H NMR (500 MHz, CDCl\textsubscript{3})\par}
\vspace{5mm}
\includegraphics[width=0.9\textwidth,height=96mm,keepaspectratio]{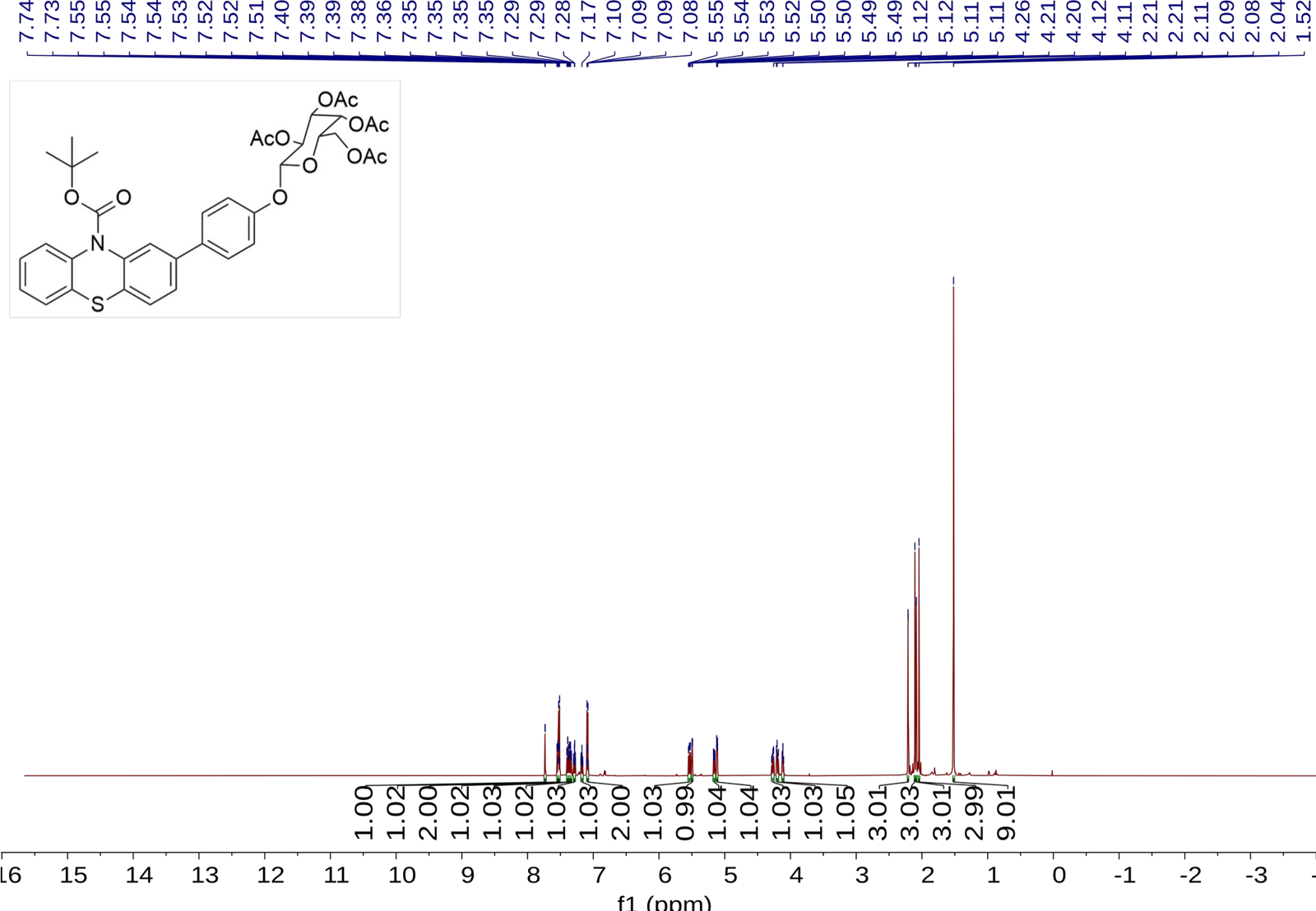}
\par\smallskip
{\small \textbf{b}, compound 34, \textsuperscript{1}H NMR (500 MHz, CDCl\textsubscript{3})\par}
\vspace{5mm}
\caption{\textbf{NMR and HRMS spectra of compound C2 and its intermediates.} Compound numbers follow Figure~\ref{fig:4}; spectra are reproduced as recorded.}
\label{fig:A-specC2}
\end{figure}
\clearpage

\begin{figure}[p]
\ContinuedFloat
\centering
\includegraphics[width=0.9\textwidth,height=96mm,keepaspectratio]{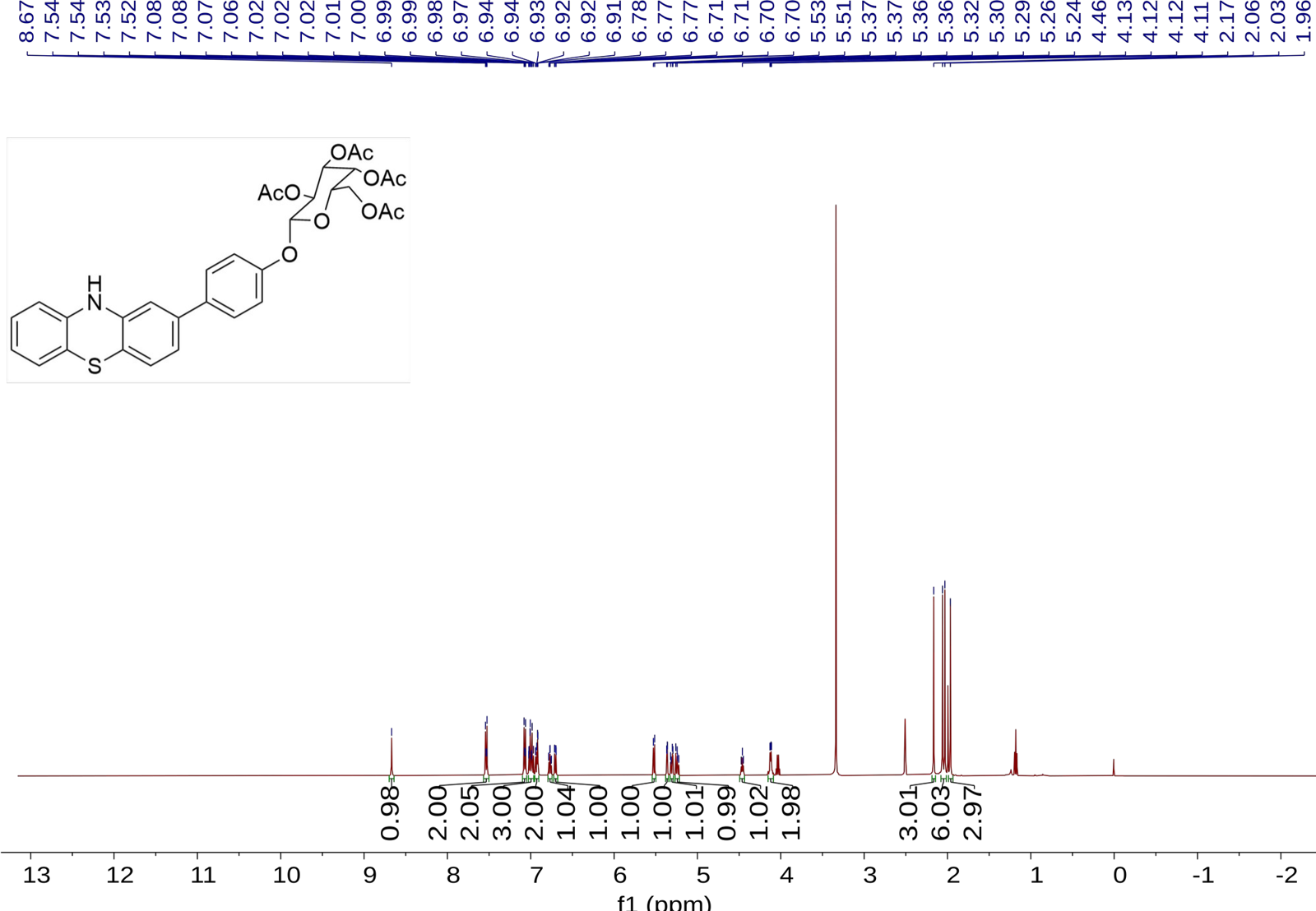}
\par\smallskip
{\small \textbf{c}, compound 28, \textsuperscript{1}H NMR (500 MHz, DMSO-\textit{d}\textsubscript{6})\par}
\vspace{5mm}
\includegraphics[width=0.9\textwidth,height=96mm,keepaspectratio]{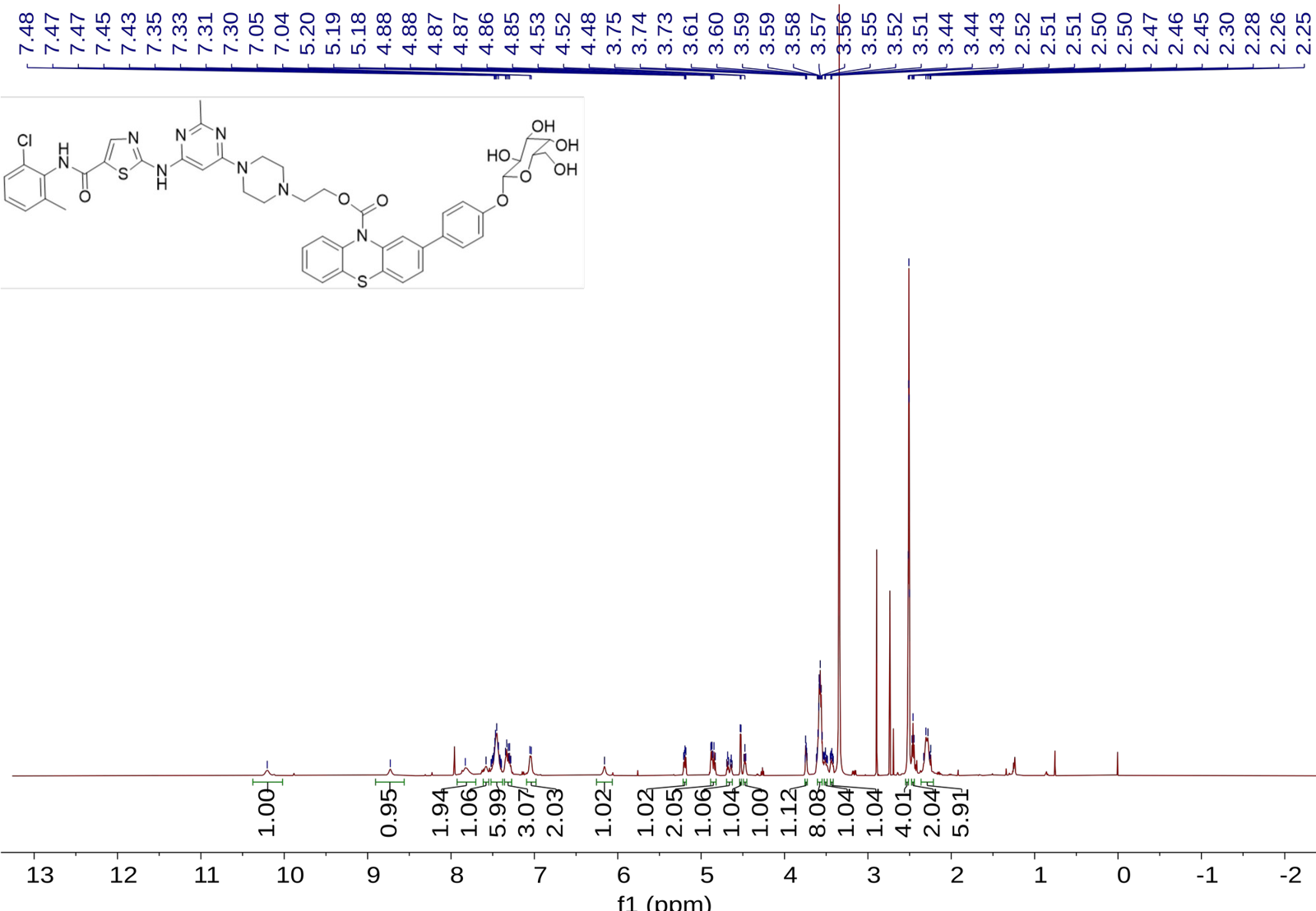}
\par\smallskip
{\small \textbf{d}, C2, \textsuperscript{1}H NMR (500 MHz, DMSO-\textit{d}\textsubscript{6})\par}
\vspace{5mm}
\caption[]{\textit{Continued.} \textbf{NMR and HRMS spectra of compound C2 and its intermediates.}}
\end{figure}
\clearpage

\begin{figure}[p]
\ContinuedFloat
\centering
\includegraphics[width=0.9\textwidth,height=96mm,keepaspectratio]{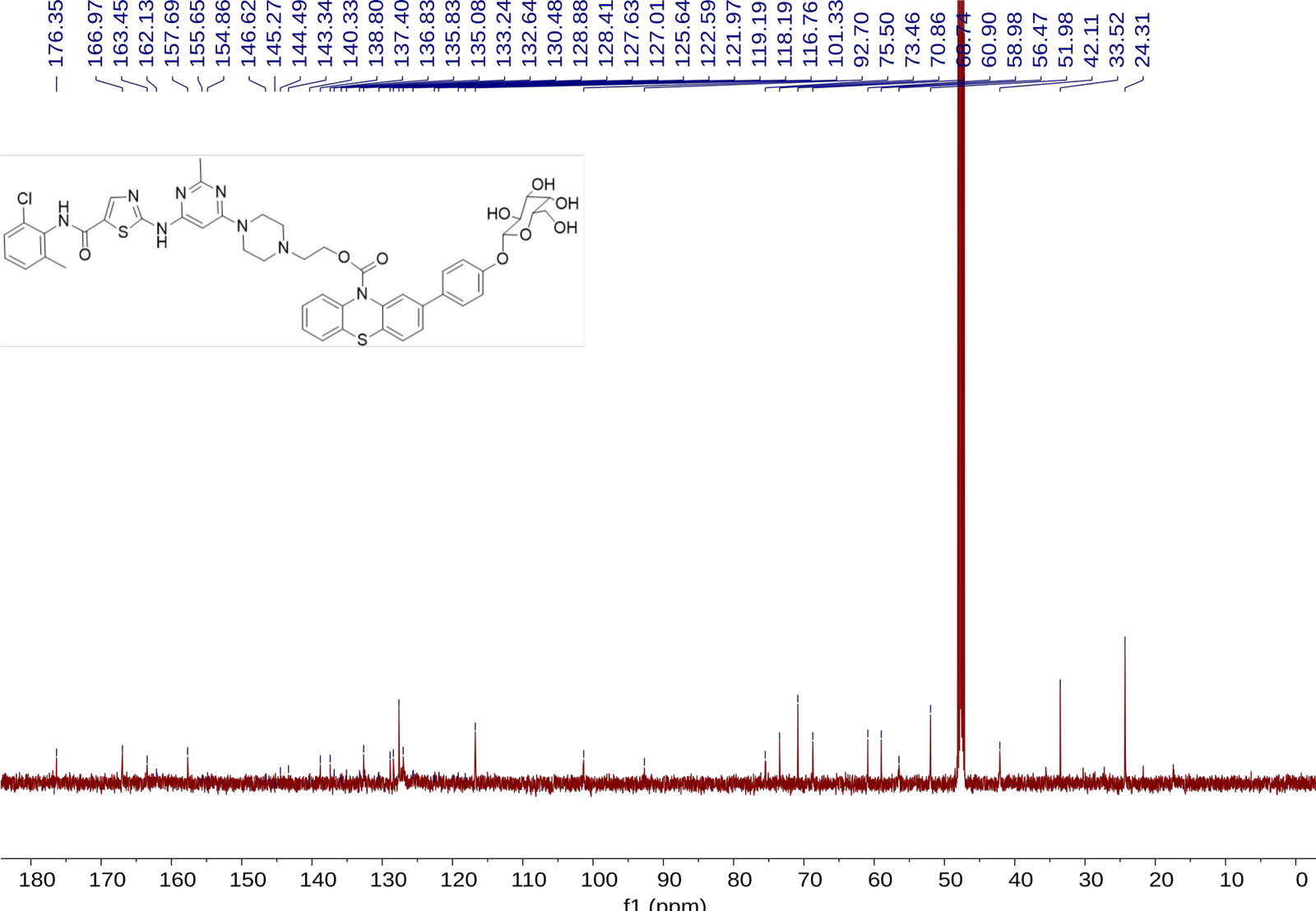}
\par\smallskip
{\small \textbf{e}, C2, \textsuperscript{13}C NMR (126 MHz, CD\textsubscript{3}OD)\par}
\vspace{5mm}
\includegraphics[width=0.9\textwidth,height=96mm,keepaspectratio]{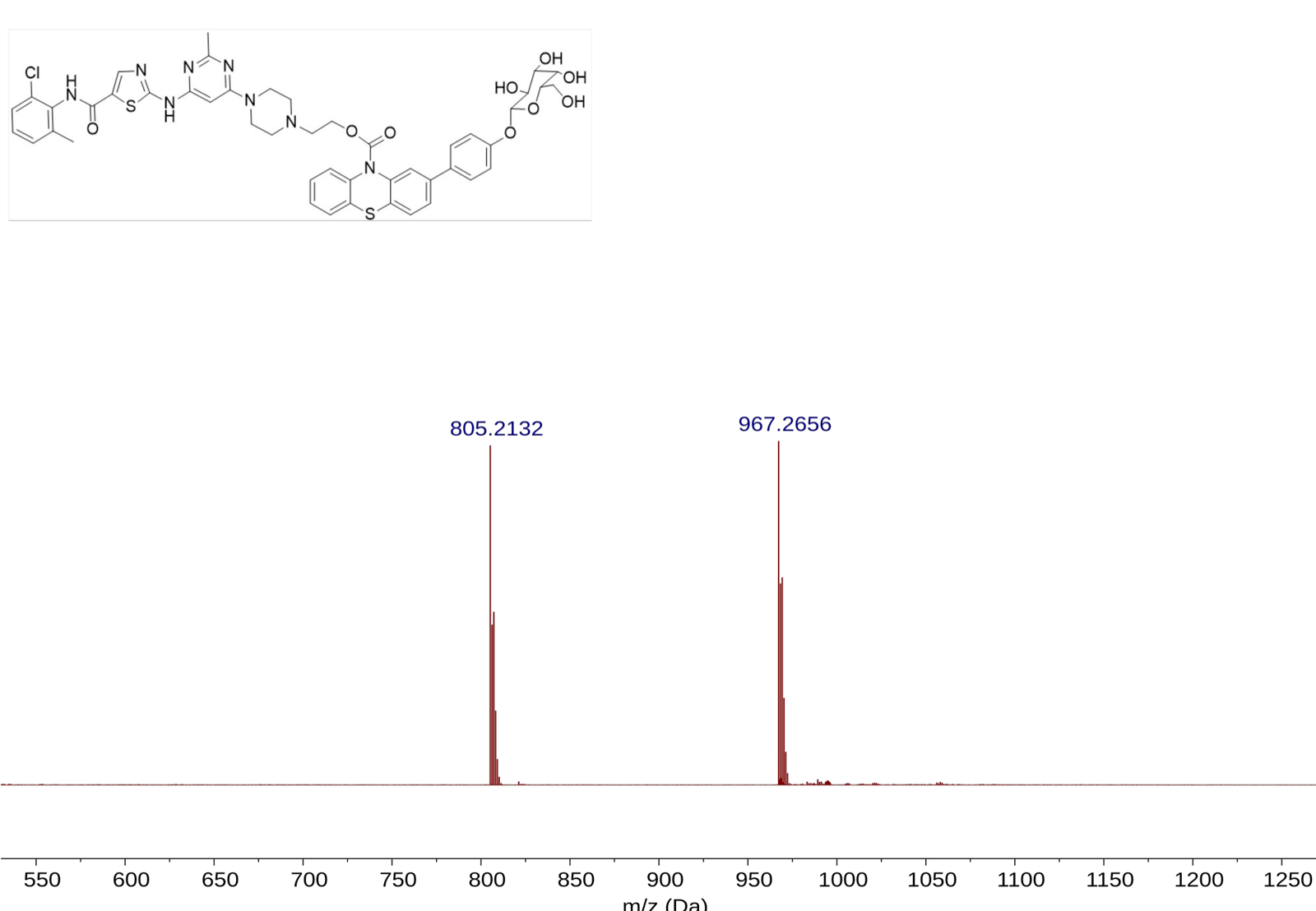}
\par\smallskip
{\small \textbf{f}, C2, HRMS\par}
\vspace{5mm}
\caption[]{\textit{Continued.} \textbf{NMR and HRMS spectra of compound C2 and its intermediates.}}
\end{figure}
\clearpage

\end{document}